\documentclass[twoside,11pt]{article}

\usepackage[preprint]{jmlr2e}
\providecommand{\texorpdfstring}[2]{#1}
\hypersetup{hidelinks,hypertexnames=false}

\usepackage{mathtools,amsfonts,amssymb}

\usepackage[T1]{fontenc}
\usepackage[utf8]{inputenc}
\usepackage{lmodern}

\usepackage{algorithm,algpseudocode}
\usepackage{textcomp}
\usepackage{gensymb}
\usepackage{url}
\usepackage{float}
\usepackage{tabularx,multirow}
\usepackage{booktabs}
\usepackage{pgfplots}
\pgfplotsset{compat=1.18}
\usetikzlibrary{patterns}
\usepackage{cleveref}
\crefname{equation}{}{}
\Crefname{equation}{}{}
\crefname{theorem}{Theorem}{Theorems}
\crefname{lemma}{Lemma}{Lemmas}
\crefname{definition}{Definition}{Definitions}
\crefname{assumption}{Assumption}{Assumptions}
\crefname{proposition}{Proposition}{Propositions}
\crefname{principle}{Principle}{Principles}

\usepackage{hyphenat}

\newcommand{\figureasset}[3]{%
  \IfFileExists{#1.pdf}{%
    \includegraphics[width=\linewidth,height=#2,keepaspectratio]{#1.pdf}%
  }{%
    \IfFileExists{#1.png}{%
      \includegraphics[width=\linewidth,height=#2,keepaspectratio]{#1.png}%
    }{%
      {\setlength{\fboxsep}{6pt}%
        \fbox{%
          \parbox[c][#2][c]{0.96\linewidth}{%
            \centering
            \small
            #3\\[0.5em]
            \texttt{\detokenize{#1}}%
          }%
        }%
      }%
    }%
  }%
}

\newcommand{\best}[1]{\textbf{#1}}
\newcommand{\score}[2]{#1{\scriptsize$\pm$#2}}

\jmlrheading{}{2026}{}{}{}{}{Weitao Li and Gong Cheng}
\ShortHeadings{Generator-Aligned Representation Interfaces}{Li and Cheng}
\firstpageno{1}

\begin{document}
\title{Generator-Aligned Representation Interfaces for Diagnostic Soft Equivariance}

\author{\name Weitao Li \email 2251984@tongji.edu.cn \\
  \addr School of Mathematical Sciences\\
  Tongji University\\
  Shanghai 200092, China
  \AND
  \name Gong Cheng \email gongch@tongji.edu.cn \\
  \addr School of Mathematical Sciences\\
  Key Laboratory of Intelligent Computing and Applications (Ministry of Education)\\
  Tongji University\\
Shanghai 200092, China}

\maketitle

\begin{abstract}%
  Exact-equivariant architectures typically encode prescribed group actions in specialized operators, which can complicate their reuse with generic backbones and across data modalities. We introduce the Generator-Aligned Representation Interface (GARI), a representation-level design principle that exposes selected transformation generators to a generic sequence backbone through aligned canonical and generator-induced views. We formalize the resulting behavior using a probe-specific soft-equivariance residual defined over declared data and transformation distributions. This framework distinguishes representation consistency from task robustness and exact equivariance, and localizes residual mismatch to interface construction, shared stream processing, and terminal fusion. We instantiate the interface as GARI-Net, which constructs generator-indexed streams, converts them into a common interaction frame, processes them with shared parameters, repairs ordering-induced context mismatch, enables cross-stream information exchange, and aggregates them using inter-stream discrepancy. Direct Equivariance Error (DEE) provides a frozen-checkpoint diagnostic of the prescribed representation relation under known token or voxel actions. Experiments on genomic sequences, images, and three-dimensional point clouds examine sequence reversal, planar rotations and reflections, and controlled axial transfer. Across these settings, the same interface principle supports task-relevant transformation consistency and generalization to declared held-out probes without requiring group-specific redesign of the sequence backbone. GARI therefore provides a portable diagnostic complement to hard-equivariant architectures: it makes generator structure accessible, learnable, and measurable, while finite-probe evidence remains distinct from certification of exact equivariance over a continuous group.
\end{abstract}


\section{Introduction}

Architectures with built-in equivariance enforce a prescribed transformation by encoding the corresponding group action in their operators. This approach underlies powerful models based on group convolutions, steerable networks, harmonic representations, and geometric tensor operations~\citep{cohen2016group,weiler2018learning,cohen2018spherical,thomas2018tensor}. Such designs can provide exact equivariance under the specified action, but guarantee typically comes with commitments to a particular group, feature geometry, or operator family. Adapting them to a new group, a generic backbone, or a different data modality can therefore require redesigning both the operator and its representation structure. At the same time, modern indexed-representation backbones offer a complementary form of portability: images, genomic sequences, and geometric data can all be serialized and processed by Transformer-like or state-space sequence models~\citep{Zhu2024ViM,hatamizadeh2024mambavision,Schiff2024Caduceus}. Yet serialization alone does not specify how a transformation of the input should induce a corresponding transformation on the resulting internal representation. This leaves an integration gap between exact-by-construction equivariant architectures and reusable cross-modal backbones. We therefore ask: how can group structure be exposed to a generic backbone without redesigning every layer as a hard equivariant operator?

We address this gap with the \textbf{Generator-Aligned Representation Interface} (GARI), a modality-agnostic interface for exposing transformation-generator structure to general-purpose backbones over indexed representations through aligned representation views. GARI uses selected generators of transformation to construct a reference view and one or more aligned transformed views of the same input. At the interface level, these views are processed with shared parameters, mapped into a common representation system, allowed to exchange information, and combined using a rule that can account for disagreement across views. We instantiate this interface as \textbf{GARI-Net}, the sequence-based model studied in this paper. GARI-Net represents the views as ordered streams and processes them with a shared sequence backbone. Rather than enforcing exact equivariance in every layer, GARI serves as a reusable interface that makes generator structure visible, trainable, and empirically falsifiable inside more generic neural models.


Moving group structure from the operator to the representation interface changes the type of equivariance claim that can be made. A hard-equivariant architecture constrains its function class so that the prescribed relation $F(\alpha_{g}x) = \tau_{g}F(x)$ holds by construction for every input $x$ and group element $g$. GARI does \emph{not} enforce this identity inside every layer. We therefore study \emph{soft equivariance}, defined by the residual with which a learned representation satisfies the same transformation relation. For a data distribution $\nu$, a declared transformation distribution $\mu$, and a representation discrepancy $\Delta$, we write
\[
  \mathcal{E}_{\mu,\nu}^2 = \mathbb{E}_{x\sim\nu, g\sim\mu}\Bigl[\Delta\bigl(F(\alpha_{g}x), \tau_{g}F(x)\bigr)^2\Bigr].
\]
Exact equivariance corresponds to a vanishing residual over the full action, whereas soft equivariance refers to a small, generally nonzero residual under a specified data and transformation distribution.

The population residual is not directly observable in a finite experiment. We estimate it on frozen checkpoints using a pre-specified finite probe set and held-out evaluation examples. At the representation level, the resulting normalized empirical residual is reported as \emph{Direct Equivariance Error} (DEE): it compares the representation of a transformed input with the prescribed transformation of the canonical representation under the known token or voxel reindexing. DEE is therefore the paper's direct finite-probe measure of representation\hyp{}level soft equivariance. We complement it with two distinct forms of evidence. Held-out task accuracy tests whether the measured representation consistency remains discriminative and useful for the task, while matched component and backbone ablations test whether the behavior is attributable to the generator-aligned interface rather than to additional streams, computation, fusion capacity, or a particular sequence block. Putting together, we measure representation-level soft equivariance through empirical DEE, evaluate its task-level consequences through held-out transformation accuracy, and test its architectural source through matched ablations. Agreement among these readouts strengthens the mechanism claim, but they play different logical roles: DEE estimates the equivariance residual, transformed accuracy measures its task-level consequence, and ablations address architectural attribution.

We apply this residual-centered protocol across sequential, planar, and three-dimensional data, with each experiment specifying the representation action, finite probe family, held-out region, and attribution control. On ImageNet-1K, GARI-Net-$C_4$ (cyclic group of order $4$) improves pure-generalization rotation accuracy over the matched serialized-vision baseline by $1.34$ percentage points ($3.00$ points on the $C_4$-specific held-out region) and reduces final $C_4$ DEE from $0.983$ to $0.831$. On MNIST, broader $C_4$ generator coverage outperforms the compute-matched $C_2$ configuration on sampled-orbit generalization; relative to the same-backbone no-generator control, GARI-Net-$C_4$ improves orbit-average accuracy by $2.61$ points and achieves the best worst-case accuracy. On GenomicBenchmarks, GARI-Net obtains the highest reversed-input average without attaining the highest normal-input average, consistent with preservation under the generator action rather than generic classifier superiority. On ModelNet40, strengthening the $x$/$y$-axis interface from $C_2$ action (rotation by $180\degree$) to $C_4$ (rotations by every $90\degree$) increases held-out $z$-axis accuracy by $8.97$ points with only a $0.33$-point change in clean accuracy, whereas an explicit $x$/$y$/$z$-axis reference adds just $0.51$ points under the same no-$z$-augmentation protocol, localizing the principal behavioral gain to transfer through the strengthened $x$/$y$ interface.

Taken together, these results position GARI as a portable, diagnostic complement to hard-equivariant architectures. Whereas specialized operators impose the prescribed transformation law by construction, GARI exposes selected generator structure to generic backbones and evaluates the learned correspondence using frozen-representation residuals, held-out transformation performance, and matched ablations. Across the declared probes and controls, the evidence supports lower empirical soft-equivariance error, improved held-out behavior, and attribution of these effects to the interface. Because these findings are finite and distribution-dependent, they neither certify exact equivariance beyond the evaluated actions and data nor equate transformation robustness with equivariance or establish superiority over specialized equivariant models.

In summary, our contributions are as follows.
\begin{enumerate}
  \item \textbf{Generator-aligned representation interfaces.} We introduce GARI as a representation\hyp{}level design primitive that exposes selected group generators through comparable canonical and generator-induced views around a generic backbone. This changes the symmetry-design unit from a group-specialized operator to a reusable interface.
  \item \textbf{Residual-based soft equivariance and finite empirical measurement.} We define soft equivariance through the residual of the prescribed representation relation and instantiate its finite-probe representation-level estimate as Direct Equivariance Error. Held-out task metrics and matched ablations are then used separately to assess discriminative usefulness and architectural attribution.
  \item \textbf{A concrete sequence-model realization.} We instantiate GARI as GARI-Net, using generator-indexed streams, shared backbone processing, frame conversion, ordering-aware context repair, cross-stream interaction, and discrepancy-aware fusion.
  \item \textbf{Cross-domain evidence and backbone portability.} We evaluate the same interface principle on sequence reversal, planar cyclic and orthogonal transformations, and controlled axial $\mathrm{SO}(3)$ transfer.
\end{enumerate}

\section{Related Work and Positioning}

\subsection{Hard-Coded Equivariant Architectures}

Most equivariant neural architectures pursue equivariance by constraining the layer operator. Group equivariant convolutional networks extend the weight-sharing principle of CNNs from translations to discrete transformation groups such as planar rotations and reflections \citep{cohen2016group}. Steerable CNNs and the e2cnn framework provide a more representation-theoretic formulation, in which feature fields transform under specified group representations and convolution kernels are restricted to steerable kernel spaces \citep{weiler2018learning,weiler2019general}. Spherical CNNs move the same idea to spherical signals and $\mathrm{SO}(3)$ correlations, using harmonic analysis and generalized Fourier transforms to obtain rotation-equivariant processing on the sphere \citep{cohen2018spherical}. These methods are foundational because they show that equivariance can be built into neural operators rather than learned only from data.

The hard-equivariance paradigm is also central in 3D and scientific machine learning. Tensor Field Networks, SE(3)-Transformers, NequIP, EGNN, and Equiformer encode geometric structure through irreducible representations, spherical harmonics, tensor products, coordinate updates, or equivariant message passing \citep{thomas2018tensor,fuchs2020se3,batzner2022e3,satorras2021en,liao2023equiformer}. These methods are powerful precisely because they restrict the computation to respect a chosen group action. The cost is that the architecture becomes coupled to a particular geometric representation and operator family. A model designed around steerable kernels, $\mathrm{SO}(3)$ irreps, molecular graphs, or point-cloud geometry is not simply a generic sequence backbone with a replaceable group interface. The survey by Fei et al.~\citeyearpar{Fei2024RotationSurvey} on rotation invariance and equivariance in 3D deep learning emphasizes that exact geometric guarantees, robustness evaluation, and integration with general-purpose deep architectures remain distinct challenges. This paper follows that distinction: the ModelNet40 protocol is a controlled generator-transfer diagnostic, not a claim of complete $\mathrm{SO}(3)$-equivariant recognition. Accordingly, GARI-Net is positioned as complementary to hard equivariant layers: the comparison is between hard-coded equivariant operators and a generator-aligned interface that can be attached to generic sequence backbones.

\subsection{Lie-Group Equivariant Layers and Geometric Attention}

Several works broaden the hard-equivariance program from specific geometric settings to more general Lie-group mechanisms. LieConv constructs equivariant convolutional layers on arbitrary continuous data for specified Lie groups, using group coordinates and local convolutional structure \citep{finzi2020generalizing}. LieTransformer extends equivariance to self-attention through Lie-group equivariant attention layers \citep{hutchinson2021lietransformer}. These methods are particularly relevant because they also address generality beyond one fixed planar symmetry. Their design, however, still modifies the layer operator so that equivariance is built into the computation itself.

GARI-Net makes a different architectural choice. It does not attempt to turn every backbone layer into a Lie-group equivariant convolution or attention map. Instead, it exposes generator-indexed views to an otherwise generic sequence backbone and asks whether task learning can organize these views into measurable soft equivariance. The group-theoretic object enters through stream construction, frame conversion, cross-stream interaction, and discrepancy-aware fusion. This distinction matters for the present paper: the target is a reusable interface principle rather than a new exact equivariant operator family.

\subsection{Approximate and Soft Equivariance}

Exact equivariance is not always the right inductive bias. Real data can contain symmetry breaking, partial observability, nuisance factors, boundary effects, or task-specific asymmetries. Approximately equivariant networks relax the strict weight-sharing constraints of equivariant architectures and show that partial symmetry bias can be preferable when the underlying system is only imperfectly symmetric \citep{wang2022approximately}. Relaxed equivariance methods also explore ways to encourage equivariance through auxiliary objectives or multitask formulations rather than by fully hard-coding the symmetry into every layer \citep{elhag2024relaxed}. Related work on approximate equivariance in reinforcement learning makes a similar point in control settings where exact symmetries may not hold globally \citep{park2024approximate}.

GARI-Net shares the premise that useful symmetry structure can be softer than exact layer-wise equivariance. Its mechanism is different. Rather than relaxing a group convolution weight-sharing scheme or adding only an equivariance loss, GARI-Net constructs a generator-aligned stream interface and evaluates the resulting behavior on declared probes. The paper's DEE metric is therefore a diagnostic of frozen representations, not a proof obligation, and held-out transformation accuracy is interpreted as task-level evidence only on specified orbit regions. This is why the manuscript repeatedly distinguishes soft equivariance from ordinary robustness and exact certification.

\subsection{Sequence Backbones and Equivariant Sequence Modeling}

The other side of the paper's positioning comes from generic sequence modeling. Vision Mamba demonstrates that bidirectional state-space sequence blocks can serve as efficient visual backbones, while MambaVision shows that hybrid Mamba-Transformer designs can be competitive for image recognition and downstream vision tasks \citep{Zhu2024ViM,hatamizadeh2024mambavision}. These models motivate treating images as sequences, but they do not by themselves specify how a geometric group action should enter the token representation or how held-out transformations should be diagnosed.

Equivariant sequence modeling provides a more direct precedent. Caduceus incorporates reverse-complement equivariance into long-range DNA sequence modeling, showing that Mamba-style sequence backbones can be adapted to domain-specific symmetry \citep{Schiff2024Caduceus}. GARI-Net extends the question from one biologically motivated sequence symmetry to a more general generator-indexed interface. In the current experiments, this interface is probed in genomic reversal, planar image rotations, sampled $\mathrm{O}(2)$ transformations, and a controlled $\mathrm{SO}(3)$ axial decomposition. These are not presented as proof of universal group learning; they are structured diagnostic settings for the same interface principle.

\subsection{Positioning of GARI-Net against State-of-the-Art Equivariant Design}

GARI-Net sits between hard equivariant geometry and generic sequence modeling. From the hard-equivariance literature, it inherits the importance of specifying a group action and a representation relation. From approximate equivariance, it inherits the idea that exact symmetry may be too rigid or too costly as a universal design target. From modern sequence modeling, it inherits a backbone class that can operate across modalities once data are tokenized. The contribution is to connect these ingredients through generator-aligned stream construction and diagnostic evaluation.

The comparison is therefore not GARI-Net versus all equivariant networks, but hard-coded equivariant operators versus a generator-aligned interface that can be attached to generic sequence backbones. In this sense, GARI-Net is an interface-level counterpart to hard equivariant architecture design: it does not enforce the group law inside every layer, but makes representation-theoretic group structure and generator accessibility trainable, measurable, and transferable across generic sequence backbones. The present manuscript reports completed diagnostic evidence and pre-specifies matched Tiny-ImageNet component attribution as the journal-version closure. This positioning keeps the paper's claim bounded: GARI-Net does not certify exact continuous-group equivariance, but it offers a trainable and testable route for exposing Lie-group generator structure to general sequence models.

\begin{table}[tbp]
  \centering
  \caption{Positioning GARI and GARI-Net against representative equivariant model families. The table summarizes mechanism, portability, scaling pressure, and claim type; it is intended as a taxonomy rather than a leaderboard.}
  \label{tab:gari-positioning}
  \scriptsize
  \setlength{\tabcolsep}{2.5pt}
  \renewcommand{\arraystretch}{1.10}
  \begin{tabularx}{\textwidth}{>{\raggedright\arraybackslash}p{0.18\textwidth}>{\raggedright\arraybackslash}p{0.24\textwidth}>{\raggedright\arraybackslash}p{0.13\textwidth}>{\raggedright\arraybackslash}p{0.23\textwidth}>{\raggedright\arraybackslash}X}
    \toprule
    Family & Mechanism & Backbone freedom & Scaling driver & Claim type \\
    \midrule
    Group CNN / G-CNN & Group lifting and group convolution & Low & Sampled group elements or orientation channels & Hard equivariance \\
    Steerable / e2cnn & Representation-constrained kernels & Low & Field types, bases, irreps & Hard / structured equivariance \\
    $\mathrm{SE}(3)$ / $\mathrm{E}(3)$ geometric networks & Irreps, tensor products, message passing & Low--medium & Neighbors, irrep degree, tensor paths & Geometric equivariance \\
    LieConv / LieTransformer & Lie-group equivariant layer operators & Medium & Group coordinates, exp/log maps, equivariant kernels or attention & General hard-equivariant layers \\
    GARI / GARI-Net & Generator-indexed sequence interface & High & Exposed generator streams and diagnostic probes & Soft, measurable, falsifiable interface \\
    \bottomrule
  \end{tabularx}
\end{table}

The scaling column should be read as an integration pressure rather than a performance ranking: GARI reduces the need to redesign the backbone operator for each group or modality, while retaining a diagnostic rather than certifying claim type.

The state-of-the-art landscape should therefore be read along two axes. Along the certification axis, hard equivariant models remain stronger because the group law is built into the layer operator. Along the portability axis, however, they often require operator redesign, feature-type commitments, or domain-specific geometric machinery. GARI deliberately chooses the second axis: it sacrifices exact layer-wise certification in exchange for backbone freedom, generator-indexed reuse, and diagnostic falsifiability across tokenized modalities. Consequently, our empirical comparisons are not intended to form a universal SOTA leaderboard across all equivariant architectures. They are designed to answer whether a generator-aligned interface improves generic sequence backbones under controlled held-out transformations and representation diagnostics.

\paragraph{Practical scaling and dataset breadth.}
Beyond exactness, a practical distinction concerns how symmetry mechanisms are deployed at scale. Hard-coded equivariant architectures provide strong guarantees, but their operator-level constraints often tie the model to a particular group action, feature type, geometric domain, or equivariant operator family. This coupling does not mean that hard-equivariant methods cannot scale; rather, it can make direct integration with modern large-scale generic backbones less straightforward, especially when the same backbone class is expected to serve images, biological sequences, and geometric tokenizations. GARI takes a different trade-off: it gives up exact layer-wise certification and instead exposes generator structure through a reusable representation interface, instantiated here as generator-indexed sequence streams. This design is naturally compatible with large-scale sequence backbones and standard large datasets, as reflected by ImageNet-1K serving as the visual anchor in our evidence chain, with GenomicBenchmarks, MNIST $\mathrm{O}(2)$, and ModelNet40 $\mathrm{SO}(3)$ probing complementary sequence, controlled generator-coverage, and held-out axial-transfer settings. Tiny-ImageNet is then reserved as the controlled scale for matched component attribution. Thus, the advantage of GARI is not stronger mathematical certification than hard-equivariant layers, but practical portability, backbone freedom, and dataset breadth under an explicitly diagnostic soft-equivariance contract.

\section{A Diagnostic Framework for Soft Equivariance}
\label{sec:diagnostic-framework}

This section defines the soft\hyp{}equivariance framework used as a diagnostic evaluation throughout the paper. Whereas hard\hyp{}equivariance architectures enforce a transformation relation through operator design, GARI asks whether exposing generator structure through a representation interface enables a generic backbone to exhibit measurable transformation consistency on specified probes. The framework proceeds from group actions and exact equivariance to a residual-based definition of soft equivariance, finite-probe measurement through Direct Equivariance Error (DEE), the role generator probes, and the localization of residual error within generator-aligned interfaces. Its purpose is empirical and falsifiable: it characterizes \emph{measurable soft equivariance under specified probes}, rather than certifying exact equivariance over a continuous group. The architectural realization is deferred to the next section.

\subsection{Exact Equivariance and Soft Equivariance Residual}

Let a group $G$ act on a data space $D$ through $\alpha:G\curvearrowright D$, and let $\tau:G\to\mathrm{GL}(\mathcal{Y})$ specify the corresponding action on a representation space $\mathcal{Y}$. A map $F:D\to\mathcal{Y}$ is \emph{exactly} $G$-equivariant if
\begin{equation}
  \label{eq:exact-equivariance}
  F(\alpha_gx)=\tau_gF(x),
  \qquad \forall x\in D,\ \forall g\in G.
\end{equation}
Hard\hyp{}equivariance architectures enforce this relation through constraints on their operators or function class. Invariance is the special case $\tau_g=I$. GARI does not impose \cref{eq:exact-equivariance}; it asks whether a generic backbone exhibits measurable transformation consistency after selected generator structure is exposed through a representation interface.

To state this question, fix a data distribution $\nu$, a finite diagnostic probe set (usually a subgroup) $G_0\subseteq G$, a declared probe distribution $\mu_{G_0}$, and a metric $d_{\mathcal{Y}}$ on $\mathcal{Y}$.
\begin{definition}[Soft equivariance residual]
  \label{def:soft-equivariance-residual}
  The soft\hyp{}equivariance residual of $F$ on $G_0$ is
  \begin{equation}
    \mathcal{E}_{G_0}(F)^2
    =\mathbb{E}_{\substack{x\sim\nu\\g\sim\mu_{G_0}}}
    \left[
      d_{\mathcal{Y}}\!\left(F(\alpha_gx),\tau_gF(x)\right)^2
    \right].
    \label{eq:soft-equivariance-residual}
  \end{equation}
  A probe $g$ may be an exposed generator, a held-out transformation, an orbit element, or a sampled transformation from a compact evaluation region.
\end{definition}
Soft equivariance is an empirical residual notion. It is not a relaxation of hard-equivariance guarantees: a small value of \cref{eq:soft-equivariance-residual} describes the declared data and probe distribution only, and does not imply that the relation holds for every input or group element.

Three evidence levels must therefore remain distinct. \textbf{Prediction-level behavior} asks whether classification remains accurate under transformation, usually with a trivial output action. This is a robustness or invariance-style endpoint. \textbf{Representation-level behavior} asks whether features transform consistently under a known representation action. \textbf{Equivariance certification} would require a guarantee over the stated domain and group action; it is not claimed here. In particular, transformed accuracy can improve without feature equivariance, while a small representation residual need not imply useful predictions.

\subsection{Finite-Probe Diagnostics and Direct Equivariance Error}

Experiments cannot evaluate every $g\in G$. Each diagnostic therefore declares a finite set
\begin{equation}
  G_0\subset G
  \label{eq:finite-probe-set}
\end{equation}
and reports behavior under its associated empirical distribution. The probes used in this paper include sequence reversal, cyclic planar rotations, reflection--rotation orbit elements, held-out transformations, and axial transfer directions in $\mathrm{SO}(3)$. These sets make the proposed mechanism falsifiable: poor residuals or task behavior on a declared probe directly weaken the corresponding claim. They provide evidence only on those probes, not certification on the surrounding continuous group.

In the completed experiments, \emph{Direct Equivariance Error} (DEE) is the representation-level implementation of \cref{eq:soft-equivariance-residual}. DEE is computed from frozen checkpoints by comparing the representation of a transformed input with the prescribed transformation of its canonical representation. The prescribed action is fixed by the diagnostic---for example, a known token or voxel reindexing---rather than fitted to make the residual small. DEE is evaluated after training and is not optimized as a loss.

DEE is interpreted jointly with task metrics. Held-out transformation accuracy tests whether representation consistency remains discriminative, while matched controls test whether any improvement is attributable to generator exposure rather than capacity or augmentation alone. Low DEE by itself is insufficient: a collapsed representation, or one that has discarded task-relevant variation through excessive invariance, may also yield a small residual. Detailed DEE normalizations and probe-specific formulas are therefore given with the corresponding experiments and in the supplementary material.

\subsection{Why Generator Probes Are Meaningful}

\paragraph{Local generator structure.}
Lie groups organize transformations locally through directions in their Lie algebra. For a direction $A$, the exponential curve $\exp(tA)$ supplies a structured family of nearby transformations rather than an arbitrary perturbation. Experiments can expose selected directions and evaluate other elements or directions as held-out probes. The paper uses this principle at several scales: $\mathbb{Z}_2$ sequence reversal, $C_2/C_4$ planar rotations, and axial $X/Y\rightarrow Z$ transfer in $\mathrm{SO}(3)$.

\paragraph{Accessibility viewpoint.}
Geometric accessibility motivates which generator directions are informative to expose or hold out. Lie closure and generator compositions can make a held-out direction structurally related to the exposed family, as in the local $X/Y\rightarrow Z$ motivation for $\mathrm{SO}(3)$. This is not a controllability theorem for the neural network. It organizes the diagnostic design; whether a trained representation uses the exposed structure remains an empirical question.

\paragraph{Connected and disconnected groups.}
For connected rotation groups, local generator flows provide natural directions along which to construct compact-region probes. They do not by themselves establish finite coverage. Disconnected actions require additional component probes: reflections in $\mathrm{O}(2)$ cannot be reached by continuous rotation flows and must therefore be tested separately.

Dense exactness provides an ideal boundary case under continuity assumptions: exact agreement on a dense subgroup would extend to the full group. Finite experiments do not meet that premise. Finite probes require explicit coverage and stability assumptions and therefore do not certify continuous-group equivariance. The dense-exactness, compact-probe, and generator-family reachability statements and proofs are collected in the supplementary material.

\subsection{Residual Localization in Generator-Aligned Interfaces}

The residual in \cref{eq:soft-equivariance-residual} identifies whether transformation consistency fails, but not where it fails. A generator-aligned interface separates three possible sources. After absorbing the relevant Lipschitz propagation factors into the terms, the total residual admits the diagnostic localization
\begin{equation}
  \delta_{\mathrm{total}}(x,g)
  \leq
  \delta_{\mathrm{interface}}(x,g)
  +\delta_{\mathrm{stream}}(x,g)
  +\delta_{\mathrm{fusion}}(x,g).
  \label{eq:main-product-space-bound}
\end{equation}
The supplementary material gives the product-space definitions and the triangle-inequality derivation. Equation~\eqref{eq:main-product-space-bound} is a localization device, not an assertion that training makes any term small.

\paragraph{Interface residual.}
Did the representation interface preserve the group action after serialization? This term captures mismatch introduced when transformed data are converted into indexed representations, including patch ordering, sequence reversal, and voxel ordering. A large interface residual means that the intended action is no longer faithfully visible to the downstream model.

\paragraph{Stream residual.}
Does the shared backbone process generator-aligned views consistently? This term captures incompatibility between the induced action and the processing of aligned views, including failures of shared sequence computation or cross-view interaction to preserve the declared correspondence.

\paragraph{Fusion residual.}
Does aggregation preserve useful generator information? This term captures mismatch introduced when multiple aligned representations are combined, including the possibility that aggregation ignores informative disagreement between views.

The decomposition does not claim that GARI-Net explicitly minimizes these terms. It localizes possible sources of soft\hyp{}equivariance error and motivates component-attribution experiments: reindexing controls target interface mismatch, interaction ablations target stream mismatch, and aggregation controls target fusion mismatch. The architecture that realizes these roles is specified in Section~4.

\subsection{Theory-to-Experiment Contract}

The framework is useful only if its mechanism can fail under declared tests. \Cref{tab:theory-prediction-map} maps each theoretical prediction to an empirical endpoint and an outcome that would weaken the proposed explanation.

\begin{table}[htbp]
  \centering
  \scriptsize
  \setlength{\tabcolsep}{2.5pt}
  \renewcommand{\arraystretch}{1.12}
  \begin{tabularx}{\textwidth}{>{\raggedright\arraybackslash}p{0.28\textwidth}>{\raggedright\arraybackslash}p{0.30\textwidth}>{\raggedright\arraybackslash}X}
    \toprule
    Theoretical prediction & Empirical endpoint & Failure mode \\
    \midrule
    Generator exposure creates measurable alignment & Held-out transformation accuracy $+$ DEE & No improvement over no-generator controls \\
    Generator coverage matters & $C_2/C_4/D_2/D_4$ comparisons & Gains explained by capacity only \\
    Interface alignment matters & No-reindex controls & Extra streams alone explain gains \\
    Stream interaction matters & No cross-attention ablation & No degradation \\
    Fusion uses disagreement information & Average-fusion control & Same performance \\
    $\mathrm{SO}(3)$ generator transfer & Held-out $Z$-axis accuracy & Gain only from explicit $Z$ exposure \\
    \bottomrule
  \end{tabularx}
  \caption{Theory-to-experiment contract for diagnostic soft equivariance. DEE is interpreted jointly with task behavior and controls; no row constitutes continuous-group certification.}
  \label{tab:theory-prediction-map}
\end{table}

A positive result in one row cannot compensate for failure in another. Accuracy without the required controls may reflect capacity or augmentation; low DEE without discriminative accuracy may reflect collapse; and held-out axial transfer remains a specified generator-transfer result rather than evidence of full $\mathrm{SO}(3)$ equivariance. This contract is the intended claim boundary for the experiments that follow.

\section{Model}

The above section defines soft equivariance as a diagnostic relation under specified generator probes. The model section describes how those probes are made visible to a generic sequence backbone. It instantiates the third level of the framework: neural realization of generator accessibility. GARI is a generator-indexed sequence architecture template whose essential abstraction is stream construction, shared processing, frame conversion, cross-stream interaction, and non-blind terminal aggregation. GARI-Net is the concrete neural instantiation used in this paper. It constructs one canonical stream and a finite set of generator-aligned streams, aligns their token orderings, processes them with a shared sequence backbone, repairs ordering-induced context mismatch, exchanges cross-stream information, and performs discrepancy-aware terminal fusion. GARI-Net is therefore not a fixed two-stream architecture. Dual-stream GARI-Net is the $q=1$ special case, while three-stream and four-stream GARI-Net instantiate the same principle for multiple independent exposed generators.

\begin{figure}[t]
  \centering
  \includegraphics[width=\textwidth]{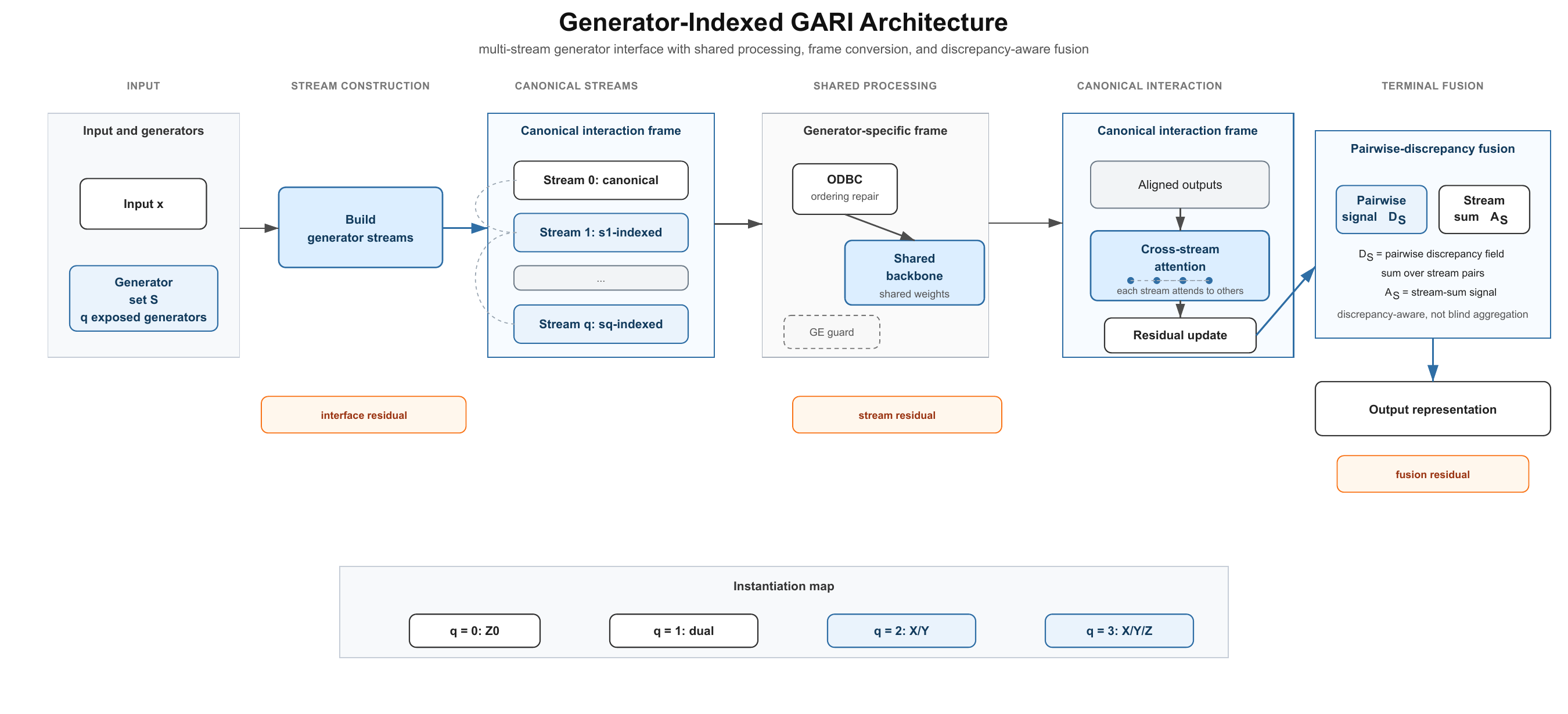}
  \caption{Generator-indexed\ GARI\ architecture\ as\ a\ neural\ realization\ of\ the representation-accessibility-realization framework. The GARI-Net instantiation used in our experiments constructs one canonical stream and $q$ generator-indexed streams from an input $x$ and a finite exposed generator set $\mathcal{S}=\{s_1,\ldots,s_q\}$. The streams are first aligned in a canonical interaction frame, then processed in generator-specific frames through ordering-aware context repair and a shared sequence backbone. After returning to the canonical frame, cross-stream attention allows each stream to exchange information with the other aligned streams, followed by a residual update. Terminal aggregation uses pairwise-discrepancy and stream-sum signals, rather than blind averaging, to produce the output representation. The orange labels localize the interface, stream, and fusion terms in the product-space decomposition; in the experiments, DEE is the reported representation-level residual diagnostic unless an internal ablation explicitly measures a stream-level quantity. The bottom row summarizes the $q=0$, $q=1$, $q=2$, and $q=3$ instantiations used as no-generator, dual-stream, $X/Y$, and $X/Y/Z$ variants.}
  \label{fig:gari-net-architecture}
\end{figure}

GARI-Net should be read as one realization of a broader generator-indexed interface template rather than as a fixed collection of architectural mechanisms. The template has five functional requirements: (i) construct canonical and generator-induced streams from a declared exposed generator family; (ii) process all streams through a shared function family so that generator consistency is meaningful; (iii) convert between generator-specific processing frames and a canonical interaction frame; (iv) provide cross-stream interaction so that aligned views can compare evidence; and (v) aggregate streams through a non-blind terminal rule that can react to inter-stream disagreement. The particular ODBC, Gradient Equilibrium, cross-stream attention, and pairwise-discrepancy fusion modules used below instantiate these requirements, but the principle is the generator-indexed interface itself. Figure~\ref{fig:gari-net-architecture} summarizes the GARI template and the GARI-Net instantiation used below: exposed generators instantiate a stream set, all streams share the same sequence backbone, and terminal aggregation is discrepancy-aware rather than a blind average.

\subsection{Generator-Indexed Stream Principle}
Let the finite exposed generator set be
\begin{equation}
  \mathcal{S}=\{s_1,\ldots,s_q\}.
\end{equation}
Each generator $s_j$ has a data-domain action $\alpha_{s_j}$ and an induced sequence-domain reindexing or token action $\rho_{s_j}$. Given a tokenization map $P_\theta$, GARI-Net constructs the product-space input
\begin{equation}
  z_{\mathcal{S}}(x)=
  \bigl(P_\theta(x),\rho_{s_1}P_\theta(x),\ldots,\rho_{s_q}P_\theta(x)\bigr),
  \label{eq:generator-indexed-streams}
\end{equation}
so the total number of streams is $q+1$. The cases used in the paper are instances of this definition: $q=0$ gives the no-generator $C_0$ (the trivial group) control with only the canonical stream; $q=1$ gives dual-stream GARI-Net; $q=2$ gives a three-stream interface such as the ModelNet40 $X/Y$ setting; and $q=3$ gives a four-stream interface such as the ModelNet40 $X/Y/Z$ explicit-generator reference. Increasing cyclic order $n$ changes the exposed generator action or orbit coverage, but it does not necessarily increase the number of independent generator streams. Adding independent generator directions increases both the stream count and the forward-pass cost.

This construction connects directly to the theory. Representation theory defines the target relation, generator accessibility chooses finite exposed generator directions, and GARI-Net exposes those directions as generator-indexed sequence streams. The product-space residual decomposition in Section~2 is naturally indexed by $\mathcal{S}$: interface mismatch arises when transformed inputs cannot be aligned to the canonical sequence frame, stream mismatch arises when the shared backbone fails to process aligned generator views consistently, and fusion mismatch arises when terminal prediction ignores cross-stream disagreement. These are diagnostic soft-equivariance locations, not exact-equivariance guarantees.

This stream principle is the architectural bridge between the mathematical objects in Section~2 and the neural implementation. Representation theory specifies the target relation through the data action $\alpha_g$ and representation action $\tau_g$. The generator-accessibility viewpoint specifies which finite directions or generator families are exposed, held out, or used as diagnostic probes. GARI-Net then turns this generator family into a trainable stream set on which a generic sequence backbone can operate. Thus the model does not hard-code the full group action into every layer; it makes selected generator evidence visible, shareable, and testable inside an ordinary sequence model.

\subsection{Stream Construction and Reindexing}
For $\mathcal{S}=\{s_1,\ldots,s_q\}$, the abstract sequence-side stream notation is
\begin{equation}
  U_0=P_\theta(x),\qquad U_j=\rho_{s_j}P_\theta(x),\quad j=1,\ldots,q.
  \label{eq:stream-construction}
\end{equation}
This equation is a notation-level description of aligned generator-indexed streams. The implementation uses
\begin{equation}
  \mathbf{T}^{(0)}=\mathrm{BuildGeneratorStreams}(x,\mathcal{S})
  =\bigl(\mathbf{T}^{(0)}_0,\ldots,\mathbf{T}^{(0)}_q\bigr),
  \label{eq:build-generator-streams}
\end{equation}
where the construction depends on the data modality. For pure sequence tasks, a generator stream can be obtained by the induced token action or reindexing $\rho_sP_\theta(x)$. For image and point-cloud tasks, a generator stream is built from a data-domain transformed input: the transformed sample is passed through the tokenizer, voxelizer, or patch embedding, and the resulting tokens are inverse-reindexed or anchor-aligned back to the canonical token grid. Thus generator streams are not merely copied canonical tokens with a permutation; they preserve a data-domain generator trace while becoming comparable in the canonical interaction frame.

ModelNet40 is one concrete point-cloud instantiation of this modality-dependent stream builder rather than a restriction of the interface template. The ModelNet40 settings instantiate this rule directly. $C_0$ uses $\mathcal{S}=\emptyset$ and is a single-stream PatchEmbed path with no generator construction. The $X/Y$ variants use $\mathcal{S}=\{R_x(180^\circ),R_y(180^\circ)\}$ for $n=2$ or $\mathcal{S}=\{R_x(90^\circ),R_y(90^\circ)\}$ for $n=4$; the implementation constructs canonical, $R_x$, and $R_y$ voxelized views, applies three pairwise Conv3d branches, averages contributions by stream, and aligns the generator streams with \texttt{inverse\_rotation\_sequence\_3d}. The $X/Y/Z$ $n=4$ explicit-generator reference uses $\mathcal{S}=\{R_x(90^\circ),R_y(90^\circ),R_z(90^\circ)\}$; the implementation constructs canonical, $R_x90$, $R_y90$, and $R_z90$ voxelized views, applies six pairwise Conv3d branches, accumulates and averages stream contributions, and aligns the streams with \texttt{inverse\_rotation\_sequence\_3d\_xyz}. The difference between $X/Y$ $n=4$ and $X/Y/Z$ $n=4$ is generator exposure, not $Z$-axis data augmentation. Both are trained without $Z$-axis rotation augmentation, so the ModelNet40 experiment separates generator exposure from training-distribution coverage.

\subsection{Generator-Indexed Building Blocks}
The following blocks are organized by interface function rather than by implementation convenience. ODBC addresses ordering-induced local-context mismatch; the shared backbone enforces a common function family; GE protects the multi-stream optimization from extreme norm imbalance; cross-stream attention supplies an interaction path between aligned generator views; and terminal fusion avoids blind averaging by using pairwise discrepancy as a deterministic disagreement signal. These blocks are therefore evaluated as functional components of the generator-indexed interface, not as independent claims of exact equivariance.
\subsubsection{One-dimensional Backward Convolution}
ODBC is applied independently to each aligned stream before the shared sequence backbone. Its role is to reduce local-context mismatch caused by generator-induced reindexing, especially when the sequence operator is directional or autoregressive. Concretely, ODBC is a one-dimensional convolutional layer along the sequence dimension. With kernel size $K$, the implementation uses right zero padding of size $K-1$, allowing a feature at each temporal location to integrate information from following positions. Thus every aligned stream receives complementary local context before entering \texttt{AutoregressiveBlock}.
Appendix~\ref{app:separable-lie-soft-equivariance} states the corresponding reversal/local-context lemma: causal windows do not commute with sequence reversal, while symmetric or bidirectional local context removes this structural mismatch away from boundary effects. ODBC is therefore a context-repair approximation for directional or autoregressive sequence operators, not a layer that enforces exact equivariance. The compatibility intuition is strongest away from sequence boundaries or under symmetric/circular padding ideals; practical zero padding can introduce boundary artifacts, which are treated as implementation-dependent effects to be checked through ablations and diagnostics.

\subsubsection{AutoregressiveBlock}
After alignment, the stream set can be written as $X\in\mathbb{R}^{(q+1)\times B\times L\times D}$. GARI-Net applies the same sequence operator to every stream,
\begin{equation}
  Y_j=f_\omega(X_j),\qquad j=0,\ldots,q.
  \label{eq:shared-sequence-backbone}
\end{equation}
Sharing $f_\omega$ keeps all streams in the same function family and prevents the model from becoming an ensemble of unrelated stream-specific networks. Mamba2, an SSM block, a Transformer-like block, or a GRU are all possible instantiations of $f_\omega$. The ModelNet40 implementation instantiates $f_\omega$ with Mamba2, while the framework itself only requires a shared sequence backbone.

\subsubsection{Gradient Equilibrium}
For $q+1$ streams, Gradient Equilibrium (GE) is a protective multi-stream norm guard. It monitors stream norms
\begin{equation}
  n_i=\|X_i\|_2+\epsilon,
  \qquad i=0,\ldots,q.
\end{equation}
It applies two safeguards: minimum-norm protection to avoid vanishing stream contributions, and pairwise norm-ratio clamping to prevent one stream from dominating shared-backbone updates or cross-stream attention.

GE is protective stabilization, not the main source of equivariance. It should be reported through trigger frequency and stream-norm-ratio diagnostics, and it is excluded from the main component-attribution claim unless separately diagnosed. Detailed implementation rules are provided in the supplementary material. When the intended representation action is norm-preserving, comparable stream norms are a useful sanity condition for small residuals, but the residual itself must still be learned by the shared backbone, alignment modules, and terminal fusion.

\subsubsection{Cross-Stream Attention}
After GE limits extreme stream-magnitude imbalance, one GARI-Net implementation of the generator-indexed interface uses leave-one-out cross-stream interaction. For $q\geq1$, each target stream $i$ receives an averaged update from all non-target streams $j\neq i$:
\begin{equation}
  \Delta_i=\frac{1}{q}\sum_{j\neq i}\operatorname{Attn}(Q_j,K_i,V_i),
  \qquad
  \tilde{Y}_i=Y_i+\Delta_i.
  \label{eq:leave-one-out-cross-stream-attention}
\end{equation}
This direction is intentional: for target stream $i$, the other streams provide query-side compatibility signals, while the target stream provides the key/value tensors, and the non-target contributions are averaged. When $q=1$, the sum has one non-target stream and the rule reduces to the dual-stream interaction between the canonical stream and one generator stream. When $q=0$, cross-stream attention is skipped and the encoder reduces to the canonical-stream control.

The attention scores learn a shared bilinear compatibility, not a distance, cosine similarity, metric, or calibrated probability. GE reduces domination by stream magnitude, but the compatibility signal remains learned and implementation-dependent. RoPE is incorporated with matched generator-induced anchors so that relative positional information does not introduce an avoidable asymmetric cue between aligned streams.

Beyond simple feature exchange, cross-stream attention plays two roles. First, it provides an explicit interaction path by which generator-indexed views can compare aligned evidence. Second, because the same projections and positional conventions are reused across streams, it avoids turning each stream into a stream-specific network. In the ideal aligned limit this does not prove exact equivariance, but it removes an obvious source of symmetry breaking while preserving global contextual exchange.

\subsection{Encoder Composition and Discrepancy-Aware Fusion}
\subsubsection{Pre-Fusion Multi-Stream Blocks}
The encoder consists of several pre-fusion generator-indexed blocks followed by one terminal fusion block. A pre-fusion block preserves the stream set rather than prematurely collapsing it. Each block applies a linear projection and normalization, maps generator-indexed streams into the generator-specific processing order by reindexing or anchoring, applies ODBC to each stream, processes every stream with the shared sequence backbone, applies GE as a stream-set norm guard, maps generator-indexed streams back to the canonical interaction frame, performs leave-one-out cross-stream attention, and updates each stream by a residual projection. This preserves canonical and generator-indexed evidence until the terminal stage.

Formally, after shared sequence processing and cross-stream interaction, the stream set remains $\{\tilde{Y}_0,\ldots,\tilde{Y}_q\}$. Keeping these streams separate supports the product-space residual interpretation: interface errors, stream-processing errors, and fusion errors remain distinguishable diagnostic locations before the final head.

\subsubsection{Pairwise-Discrepancy Terminal Fusion}
The terminal fusion block collapses the generator-indexed stream set into a single task representation. It is discrepancy-aware rather than a blind stream average. Given aligned post-interaction streams $\tilde{Y}_0,\ldots,\tilde{Y}_q$, GARI-Net computes the pairwise disagreement field
\begin{equation}
  D_{\mathcal{S}}=\sum_{0\leq i<j\leq q}(\tilde{Y}_i-\tilde{Y}_j)^2,
  \label{eq:pairwise-discrepancy}
\end{equation}
and the stream-sum representation
\begin{equation}
  A_{\mathcal{S}}=\sum_{i=0}^{q}\tilde{Y}_i.
  \label{eq:stream-sum-representation}
\end{equation}
The negative discrepancy
\begin{equation}
  R_{\mathcal{S}}=-D_{\mathcal{S}}
  \label{eq:negative-discrepancy-signal}
\end{equation}
acts as a deterministic disagreement signal. Terminal fusion first projects the normalized discrepancy and stream-sum signals into query/key/value tensors,
\begin{align}
  (Q_{\mathcal{S}},K_{\mathcal{S}})
  &=\operatorname{Split}\!\left(\operatorname{LN}(R_{\mathcal{S}})W_{qk}\right),
  \qquad
  V_{\mathcal{S}}=\operatorname{LN}(A_{\mathcal{S}})W_v,
  \label{eq:discrepancy-aware-qkv}
\end{align}
and then forms the fused representation as
\begin{equation}
  Y_{\mathrm{fuse}}
  =\operatorname{Attn}\!\left(Q_{\mathcal{S}},K_{\mathcal{S}},V_{\mathcal{S}}\right)
  +\operatorname{LN}(A_{\mathcal{S}}).
  \label{eq:discrepancy-aware-fusion}
\end{equation}
The fused representation is then passed through a Linear/GELU/DropPath-style projection and combined with the canonical residual before the task head. This mechanism uses pairwise stream disagreement as the Q/K signal and the stream-sum representation as the V signal, so terminal fusion attends using generator-disagreement structure without treating streams as a blind average. It is deterministic discrepancy-aware terminal fusion rather than calibrated probabilistic stream weighting.

For $q=0$, the discrepancy sum is empty and the block degenerates to a canonical-stream terminal projection. Detailed fusion pseudocode is provided in the supplementary material.

\subsubsection{Overall Encoder Implementation}
The main encoder algorithm records the generator-indexed interface at the level needed for the framework: stream construction, directional frame conversion, shared processing, protective stabilization, cross-stream interaction, and discrepancy-aware aggregation. The frame-conversion operators are directional: generator-frame conversion temporarily moves canonical-aligned generator streams into the generator-specific processing order, while canonical-frame conversion returns them to the canonical interaction frame for cross-stream interaction and fusion. Anchor-aware variants for vision or CLS-style reference tokens follow the same interface roles and are detailed in the supplementary material.

Generator streams are canonical-aligned at block input, temporarily moved to the generator-specific processing frame for ordering-aware sequence processing, and then returned to the canonical interaction frame for cross-stream interaction and fusion.
\begin{algorithm}[H]
  \caption{Generator-Indexed GARI-Net Encoder}
  \label{alg:gari-net-encoder}
  \begin{algorithmic}[1]
    \Require Input $\mathbf{x}$; exposed generator set $\mathcal{S}=\{s_1,\ldots,s_q\}$; $\mathcal{L}$ blocks
    \Ensure $\mathbf{x}_{\mathrm{out}} \in \mathbb{R}^{B \times L \times D}$
    \State $\mathbf{T}^{(0)}\leftarrow\text{BuildGeneratorStreams}(\mathbf{x},\mathcal{S})$
    \For{$k = 1$ \textbf{to} $\mathcal{L}-1$}
    \State Normalize the current stream set $\mathbf{T}^{(k-1)}$ and project each stream
    \State Convert generator streams to generator-specific processing frames
    \State Apply ordering-aware context repair to each stream
    \State Apply a shared sequence backbone over the stream set
    \State Apply optional protective GE over the stream set
    \State Convert generator streams back to the canonical interaction frame
    \State Apply cross-stream interaction among aligned generator views
    \State Update each stream by residual projection to obtain $\mathbf{T}^{(k)}$
    \EndFor
    \State $\mathbf{x}_{\mathrm{out}} \leftarrow \text{PairwiseDiscrepancyFusion}(\mathbf{T}^{(\mathcal{L}-1)})$
    \State \Return $\mathbf{x}_{\mathrm{out}}$
  \end{algorithmic}
\end{algorithm}

Detailed module-level pseudocode, including ODBC, shared stream processing, Gradient Equilibrium, leave-one-out cross-stream attention, pairwise-discrepancy fusion, and the anchor-aware Vision GARI-Net specialization, is provided in the supplementary material.

The resulting encoder should be interpreted at two levels. At the implementation level, this paper uses the specific GARI-Net blocks described above. At the principle level, the essential object is the generator-indexed stream interface: a finite exposed generator set $\mathcal{S}$, a modality-dependent stream builder, shared sequence processing, directional frame conversion, cross-stream interaction, and discrepancy-aware aggregation. The experiments test this principle through its concrete GARI-Net realization. Other sequence backbones or alternative modules could instantiate the same template, provided they preserve these interface roles and are evaluated by the same held-out and residual diagnostics.

This algorithm describes how GARI-Net exposes finite generator probes to a shared sequence backbone. It does not imply exact equivariance, full group certification, or neural controllability. The same algorithm yields $C_0$, dual-stream, $X/Y$ three-stream, and $X/Y/Z$ four-stream variants by changing the generator set $\mathcal{S}$. The empirical sections determine whether the resulting interface improves held-out task robustness and DEE diagnostics.

\subsubsection{Complexity and Scaling with Generators}
With $q$ exposed generators, GARI-Net uses $q+1$ streams. Backbone parameters are shared, so parameter growth is small and mostly module-dependent, while forward compute scales approximately with the number of streams; cross-stream interaction and pairwise fusion add implementation-dependent overhead. Cyclic order and independent generator count should be distinguished: increasing cyclic order can change generator views or probe granularity, whereas adding independent generator directions increases the stream set. Thus GARI-Net should be evaluated with matched controls and ablations; the claim is a generator-indexed interface for diagnostic soft equivariance, not free robustness or an efficiency guarantee.

\begin{table}[tbp]
  \centering
  \caption{Model instantiations of the generator-indexed GARI-Net interface. Stream count is $q+1$. Cyclic order controls the exposed generator action; independent generator directions control the number of streams. The $X/Y/Z$ $n=4$ setting exposes the $Z$ generator structurally but uses no $Z$-axis rotation augmentation.}
  \label{tab:model-instantiations}
  \scriptsize
  \setlength{\tabcolsep}{3pt}
  \renewcommand{\arraystretch}{1.10}
  \begin{tabularx}{\textwidth}{>{\raggedright\arraybackslash}p{0.18\textwidth}>{\raggedright\arraybackslash}p{0.24\textwidth}c>{\raggedright\arraybackslash}p{0.18\textwidth}>{\raggedright\arraybackslash}X}
    \toprule
    Variant & Generator set $\mathcal{S}$ & Streams & Exposed action & Role \\
    \midrule
    $C_0$ / $n=0$ & $\emptyset$ & 1 & None & No-gen control \\
    Dual-stream GARI-Net & $\{s\}$ & 2 & One generator & Minimal interface \\
    Image/MNIST $C_2$/$C_4$ & One cyclic generator & 2 & $C_2$ or $C_4$ & Cyclic probe \\
    ModelNet40 $X/Y$ $n=2$ & $\{R_x(180^\circ),R_y(180^\circ)\}$ & 3 & $C_2$ on $X/Y$ & Weak X/Y baseline \\
    ModelNet40 $X/Y$ $n=4$ & $\{R_x(90^\circ),R_y(90^\circ)\}$ & 3 & $C_4$ on $X/Y$ & Held-out Z transfer \\
    ModelNet40 $X/Y/Z$ $n=4$ & $\{R_x(90^\circ),R_y(90^\circ),R_z(90^\circ)\}$ & 4 & $C_4$ on $X/Y/Z$ & Explicit Z ref. \\
    \bottomrule
  \end{tabularx}
\end{table}

\section{Experiments}
The experiments are organized around a soft-equivariance design philosophy. Rather than hard-coding every desired group action into the network, we expose generator-accessibility interfaces through the generator-indexed GARI-Net architecture and test whether task supervision can make the corresponding equivariant behavior emerge. The section therefore follows a diagnostic chain rather than a list of independent benchmarks: we first monitor whether a sequence-level symmetry signal appears under ordinary vision supervision, then verify the sequence-side premise, and finally test whether strengthened generator interfaces induce vision-level robustness over increasingly rich symmetry groups.

Concretely, the first part studies the reverse direction, from vision supervision back to a sequence-level symmetry diagnostic. The second part follows the theoretical direction, from sequence equivariance to induced planar vision equivariance: it begins with a sequence-reversal diagnostic, then moves to planar cyclic rotations and planar orthogonal symmetries. The final part isolates the three-dimensional case as a held-out generator-transfer probe under $\mathrm{SO}(3)$ decomposition. This organization keeps the central claim aligned with the design principle: GARI-Net is not presented as a hard exact-equivariant constraint, but as a learnable architecture that makes generator structure visible, trainable, and transferable through data.

The experiments below form a diagnostic evidence chain rather than a single benchmark claim. The reported diagnostic experiments cover ImageNet-1K, GenomicBenchmarks, MNIST $\mathrm{O}(2)$, and ModelNet40 controlled $\mathrm{SO}(3)$ held-out generator transfer. ImageNet-1K provides the large-scale visual anchor, GenomicBenchmarks tests the sequence-reversal premise, MNIST $\mathrm{O}(2)$ provides a controlled generator-coverage probe, and ModelNet40 tests held-out generator transfer under an $\mathrm{SO}(3)$ axial decomposition. These experiments test different consequences of the same generator-aligned interface principle: held-out task robustness, orbit-localized behavior, and representation-level DEE.

Comparison policy. We separate three kinds of empirical comparisons. First, mechanism-matched baselines test whether adding generator-aligned structure improves a comparable generic sequence backbone. Second, state-of-the-art family comparisons provide context for where the method sits relative to hard equivariant, approximate equivariant, sequence-equivariant, and geometric models. Third, matched component attribution tests whether the gains come from generator-indexed alignment rather than extra streams, fusion capacity, or training dynamics. The first type is the primary evidence in the present manuscript; the second type prevents mispositioning; the third type is pre-specified through the Tiny-ImageNet attribution protocol as the journal-version attribution closure.

To make the mechanism tests falsifiable rather than benchmark-accumulative, each experiment follows the same argumentative structure: we state the mechanism claim, identify the control or matched comparison, specify the primary endpoint before interpreting secondary summaries, name the main negative or alternative explanation, state what result would weaken the claim, and then report the observed pattern. This organization is especially important for the soft-equivariance setting, where the architecture is expected to encourage group-consistent behavior rather than enforce exact equivariance by construction.

For the geometric robustness experiments, we pair the task-level endpoint with a post-hoc feature-level Direct Equivariance Error (DEE) diagnostic. ImageNet-1K asks whether final patch-token DEE aligns with planar $C_2/C_4$ rotation robustness, MNIST asks whether generator coverage changes final representation consistency under sampled $\mathrm{O}(2)$ transformations within a same-backbone control, and ModelNet40 asks whether strengthening only the $X/Y$ generators improves held-out $Z$-axis task robustness, with DEE reported as a conservative feature-level diagnostic. Across the completed geometric experiments, DEE is the reported residual diagnostic: it measures representation-level consistency on frozen checkpoints and is interpreted together with task-level robustness rather than as a separate architectural stream-residual measurement. In these completed diagnostics, DEE is computed on frozen checkpoints and is not used as a training objective, so it complements the held-out task metrics rather than replacing them.

Across the rotation experiments, we therefore treat held-out transformation generalization as the primary endpoint. Averages over all evaluated angles are reported only as diluted controls: they mix transformations already supported by the training distribution or the clean orientation with genuinely unseen transformations, and therefore understate the orbit-specific effect that GARI-Net is designed to expose. This distinction matters for GARI-Net: improving accuracy on unseen group elements is a harder extrapolation problem than improving performance on canonical or augmentation-supported inputs. Moreover, many standard ways of improving the latter are orthogonal to GARI-Net in the architectural sense: they can change the base classifier's absolute accuracy without specifying how the rotation group should act on serialized patches or how canonical and generator-indexed streams should be aligned. Examples include replacing the flat serialized backbone with a stronger hierarchical feature pyramid, increasing width/depth or input resolution, using large-scale supervised or self-supervised pretraining, adding knowledge distillation, extending the optimization schedule, or strengthening generic augmentations such as Mixup, CutMix, RandAugment, and RandomErasing~\cite{He2016ResNet,Liu2021Swin,Tan2019EfficientNet,He2022MAE,Hinton2015Distilling,Zhang2018Mixup,Yun2019CutMix,Cubuk2020RandAugment,Zhong2020RandomErasing}. These techniques are therefore complementary to the generator-aligned interface rather than substitutes for it. Our claims focus on within-protocol gains on held-out or generator-localized transformations, not on replacing those complementary accuracy-improvement techniques.

Because the experiments span images, biological sequences, and point clouds, the autoregressive core and external baselines necessarily differ across datasets. This should not be interpreted as selecting the most favorable block independently for each result. Table~\ref{tab:experiment-summary} should instead be read as an evidence-chain map: each experiment specifies its role in the theoretical validation sequence, the generator-accessibility probe being tested, and the controlled factor that supports the corresponding inference. Thus, cross-task evidence is used to test whether the same generator-aligned principle appears across modalities; quantitative claims are made within each row and through the progression of the chain, not by directly comparing absolute accuracies across rows.

\begin{table}[htbp]
  \centering
  \small
  \setlength{\tabcolsep}{2pt}
  \renewcommand{\arraystretch}{1.08}
  \begin{tabularx}{\textwidth}{>{\raggedright\arraybackslash}p{0.18\textwidth}>{\raggedright\arraybackslash}p{0.23\textwidth}>{\raggedright\arraybackslash}p{0.27\textwidth}>{\raggedright\arraybackslash}X}
    \toprule
    Evidence layer & Main endpoint & Diagnostic / control & Claim boundary \\
    \midrule
    Generator exposure & No-, weak-, and stronger-generator settings & Same backbone or matched recipe & Exposure is a probe, not exact equivariance \\
    Held-out transfer & Rotation/orbit accuracy outside supported transformations & Pre-specified held-out or generator-localized regions & Robustness alone does not certify feature equivariance \\
    DEE & Frozen-checkpoint representation residuals & Known token/voxel reindexing; no fitted action unless reported & Low DEE is supportive but not sufficient \\
    Claim boundary & Cross-experiment synthesis of positive and weakening evidence & Failure modes stated before interpretation & Finite probes do not certify a continuous group \\
    \bottomrule
  \end{tabularx}
  \caption{Compact evidence chain for the generator-aligned soft-equivariance claim. Completed rows report diagnostic evidence from held-out task robustness, orbit-localized accuracy, and frozen-checkpoint DEE.}
  \label{tab:experiment-summary}
\end{table}

\subsection{Emergent Vision-to-Sequence Equivariance}
Before testing geometric robustness, we first ask whether ordinary vision training can reduce a sequence-level generator discrepancy without explicitly optimizing that discrepancy. The purpose of this experiment is diagnostic rather than proof-oriented. We reconfigure parameter sharing within the AutoregressiveBlock and instantiate the component as the MambaVision Mixer, using it as a strong representative sequence module. The hypothesis is that a high-capacity visual sequence backbone can partially align the canonical stream and the exposed generator stream from visual supervision alone, even when no exact equivariance constraint is imposed.

Here, the MambaVision Mixer should be understood purely as a representative instantiation for empirical study, not as a restriction of the proposed framework. The objective is to characterize how the shared generator-indexed meta-architecture interacts with the generator-consistency learning ability of the underlying sequence model, rather than to claim a theorem specific to MambaVision.

In the terminology of the Methodology, this experiment monitors a training-dynamics symmetry diagnostic associated with the finite generator interface. The concrete instantiation is the $C_2$ case because it yields the cleanest controllable diagnosis inside the MambaVision Mixer: on the sequence side, the generator action reduces to sequence reversal, while on the vision side it reduces to central symmetry (a $180^\circ$ rotation). Regarding the SSM within the MambaVision Mixer, we restrict parameter sharing to the $dt\_proj$ parameter across the $q=1$ stream pair and introduce Symmetry Loss to monitor the evolution of generator-level consistency during training. Although this loss could be used as an auxiliary regularizer, we intentionally exclude it from the optimization objective. Directly penalizing this monitored diagnostic could artificially constrain the optimization landscape, thereby confounding the question of whether image-level supervision alone can induce partial generator alignment.
The diagnostic used in this section is the symmetry loss $\ell_{sym}$, a streamwise SSM-influence discrepancy defined in Appendix~\ref{app:symmetry-loss-diagnostic}. It measures whether the canonical stream and the exposed generator stream induce similar local input--output influence after the $\mathbb{Z}_2$ alignment. In the present experiment it is used only as a post-hoc monitoring signal: it is not included in the training objective, and no flipped or generator-transformed ImageNet samples are imposed during training. The claim is therefore deliberately limited. If ordinary vision supervision can partially align this $q=1$ stream pair, $\ell_{cls}$ and $\ell_{sym}$ should co-decrease during early feature formation; if exact equivariance were being enforced by the objective, the symmetry diagnostic would be expected to keep improving monotonically throughout training. The negative explanation we guard against is direct symmetry supervision, which is absent here. A result that would weaken the claim would be either no reproducible early co-decrease or a trajectory explainable by hidden generator-transformed samples. The observed pattern is an early co-decrease followed by a later plateau, which supports partial stream alignment from ordinary vision supervision but not exact $\mathbb{Z}_2$ equivariance. The EMA-curve readout makes this pattern quantitative: from epoch $1$ to epoch $50$, $\ell_{cls}$ decreases from approximately $6.94$ to $5.14$ ($-25.9\%$), while $\ell_{sym}$ decreases from $3.46\times10^{-3}$ to $1.33\times10^{-3}$ ($-61.5\%$). Over epochs $50$--$250$, $\ell_{sym}$ remains in a low band with EMA mean $1.44\times10^{-3}$ and temporal standard deviation $0.14\times10^{-3}$; the plotted three-seed min--max envelope has mean width $0.93\times10^{-3}$. Thus the diagnostic supports early co-decrease and later saturation with mild drift, rather than continued enforcement of exact equivariance.

\begin{figure}[tbp]
  \centering
  \includegraphics[width=0.95\textwidth,height=0.48\textheight,keepaspectratio]{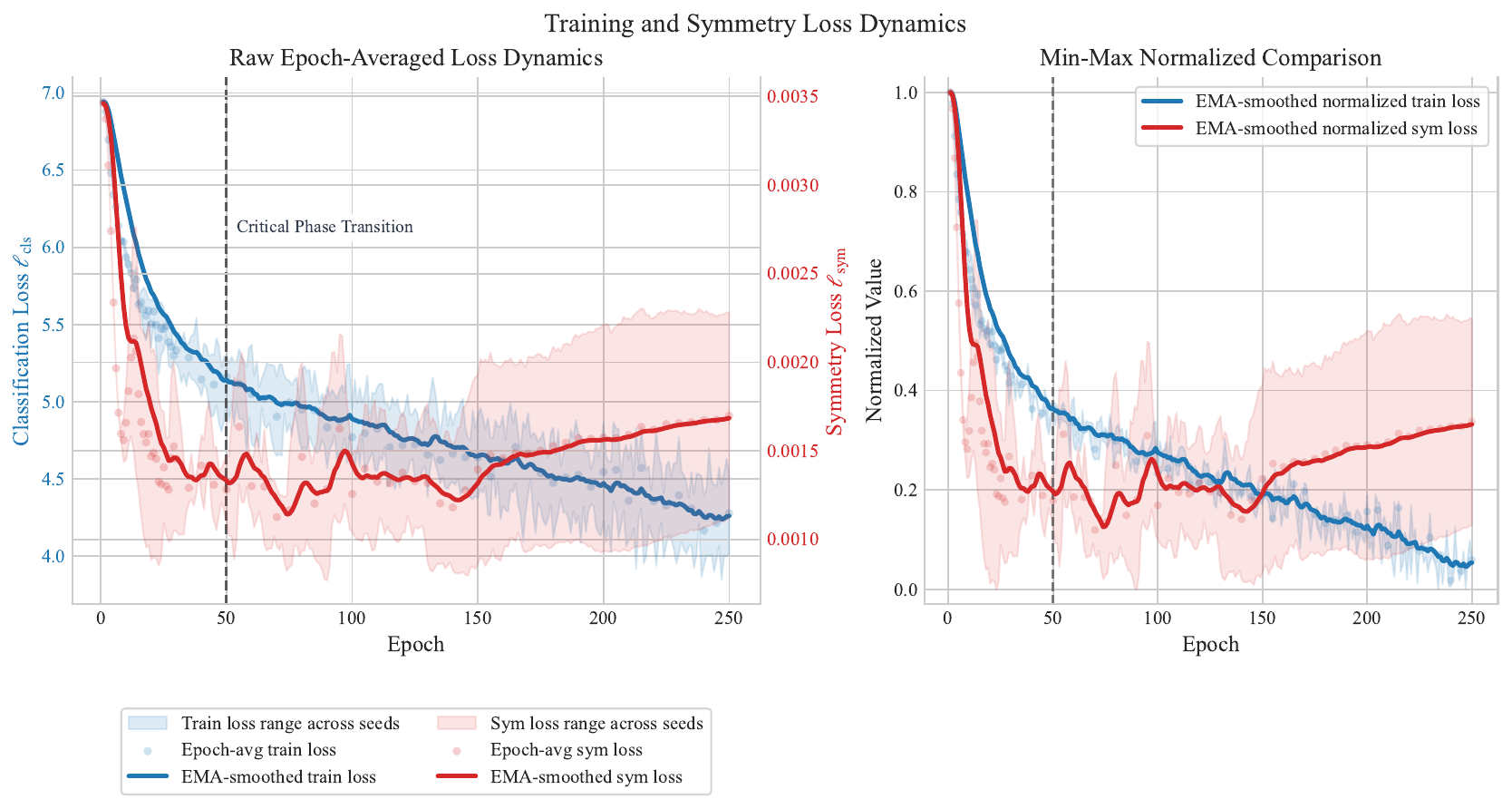}
  \caption{Analysis of loss dynamics during training. The normalized classification loss and symmetry loss decrease synchronously during the initial 50 epochs, indicating that early feature formation for the classification task is accompanied by reduced inter-stream symmetry discrepancy. Since flipped or generator-transformed samples are not imposed during training and the symmetry loss is used only for monitoring, the later plateau indicates that continued classification-loss reduction no longer requires a corresponding monotonic decrease in the $\mathbb{Z}_2$ symmetry discrepancy.}
  \label{fig:loss_dynamics_full}
\end{figure}

We conducted this diagnostic on ImageNet-1K using eight NVIDIA L40 GPUs and three independent random seeds. Given the cost of full ImageNet-1K training, this replication protocol is used to assess trial-level reproducibility of the trajectory rather than to support a high-precision estimate of a single endpoint metric. The statistical unit is the independent training run, and the claim is limited to qualitative curve behavior visible across runs. The resulting loss dynamics are plotted in Fig.~\ref{fig:loss_dynamics_full}.
\subsubsection{Synchronized Evolution Analysis}
The experimental results demonstrate a clear early synchronization between the classification objective and the symmetry diagnostic. As illustrated in the Min-Max normalized comparison, the symmetry loss $\ell_{sym}$ exhibits a sharp decay alongside the classification loss $\ell_{cls}$ during the initial 50 epochs. Because the training samples are not explicitly transformed by sequence reversal or image-level central symmetry, this alignment should not be interpreted as direct supervision for $\mathbb{Z}_2$ equivariance. A more appropriate interpretation is that the early stage of ordinary classification training learns generic low- and mid-level visual structures, such as local texture and shape cues, that are useful across both the canonical and exposed generator streams. Under the proposed parameter-sharing generator-indexed architecture, instantiated here by the $q=1$ $\mathbb{Z}_2$ stream pair, this shared feature formation naturally reduces part of the discrepancy measured by $\ell_{sym}$. Thus, the early synchronized decline reflects a general regularity of classification-driven representation learning rather than proof that the classification objective alone enforces exact equivariance.
\subsubsection{Plateau under Non-Flipped Training}
Around Epoch 50, the two trajectories begin to separate: $\ell_{sym}$ stabilizes into a regime of minor oscillations, while the classification loss continues its steady descent. This plateau is a natural consequence of the non-flipped training protocol. Once the early representation-formation stage has produced features sufficient to support the main classification objective, the remaining reduction in $\ell_{cls}$ can be achieved by class-discriminative refinements that are not required to further improve the $\mathbb{Z}_2$ symmetry discrepancy. In other words, because no flipped samples are imposed and the objective contains no explicit symmetry regularizer, late-stage classification optimization has only weak pressure to keep decreasing $\ell_{sym}$. The plateau should therefore be read as a decoupling between continued classification improvement and the monitored symmetry diagnostic, not as evidence that equivariance must keep improving throughout training.
\subsubsection{Statistical Stability and Consistency}
The narrow variance envelope across independent trials (represented by the shaded regions) underscores the numerical stability of this diagnostic pattern. Despite varying random seeds, the early co-decrease and subsequent plateau of $\ell_{sym}$ remain highly consistent. Because this ImageNet-scale diagnostic is based on three runs, we do not use it for formal hypothesis testing or for fine-grained ranking of endpoint accuracies. Instead, the evidence is the reproducible shape of the coupled trajectories: the same early alignment and later decoupling occur across the independent runs. This reproducibility suggests that the observed relationship between classification training and the symmetry measure is a stable consequence of combining ordinary classification supervision with the $q=1$ generator-indexed architecture under the present non-flipped training protocol. At the same time, the plateau reinforces the more cautious interpretation: the classification task can initially reduce inter-stream discrepancy as part of general feature formation, but it does not by itself provide a persistent objective-level requirement for exact $\mathbb{Z}_2$ equivariance.
\subsection{Sequence Premise and Planar Vision Robustness}
Having used the preceding experiment to visualize how an inter-stream symmetry diagnostic co-evolves with ordinary classification training, we next shift the focus from observation to intervention. Specifically, the following experiments investigate how the proposed meta-architecture, which is designed to promote group-equivariance learning in sequence models, can be used to reveal the influence of improved sequence-level equivariance on the equivariance of the corresponding induced planar vision model. We instantiate this meta-architecture with different autoregressive blocks and different planar generator representations, and evaluate the resulting models across sequence and two-dimensional vision tasks. These studies serve two purposes: they validate the theoretical connection established above and demonstrate the application potential of the proposed meta-architecture. Before analyzing how sequence-side robustness affects visual-task performance, however, we first need to test whether the meta-architecture helps preserve task-relevant signals under the sequence-reversal action; only then can its effect on the corresponding vision model be meaningfully assessed.

Accordingly, the subsection proceeds through increasingly targeted tests. \textit{Sequence-Side Premise: Reversal Robustness} first establishes the sequence-side premise. \textit{Cyclic-Generator Probes of $\mathrm{SO}(2)$ Rotation Robustness} then tests whether finite planar rotation generators can induce empirical two-dimensional rotational robustness. \textit{Generator-Count Effects on $\mathrm{O}(2)$ Equivariance} further studies whether increasing generator coverage, including the zero-generator baseline, strengthens equivariance over the larger planar symmetry group.
The subsequent subsection separates the qualitatively different $\mathrm{SO}(3)$ setting as the final three-dimensional test.

\subsubsection{Sequence-Side Premise: Reversal Robustness}
To verify the sequence-side premise required by the subsequent sequence-to-vision analysis, we instantiate the nontrivial generator action in the $\mathbb{Z}_2$ case as sequence reversal. This experiment does not claim a new pretrained genomic benchmark result; instead, it uses sequence-reversal robustness as an empirical proxy for sequence-side equivariance learning. We train all models from random initialization using only the standard, non-reversed training split of the eight-task GenomicBenchmarks suite~\cite{Gresova2023GenomicBenchmarks}. At test time, each checkpoint is evaluated on both the standard test set and a flipped test set obtained by reversing the input order, i.e., $x_1,\ldots,x_L \mapsto x_L,\ldots,x_1$. Since this operation only reorders the same DNA bases and no additional directional annotation is introduced by the benchmark preprocessing, the class label is preserved. This transformation is not a reverse-complement operation, and thus isolates sequence-order reversal rather than biological strand complementarity.

We compare the proposed GARI-Net sequence encoder with Caduceus-PH and Caduceus-PS. In the genomic setting, Caduceus provides a domain-specialized sequence-equivariant reference point rather than merely a generic baseline, because it explicitly incorporates reverse-complement symmetry into Mamba-style long-range sequence modeling and reports state-of-the-art-level downstream genomic performance, including on GenomicBenchmarks~\cite{Schiff2024Caduceus,Gresova2023GenomicBenchmarks}. We therefore use it as SOTA context for sequence-side symmetry modeling, while keeping the primary endpoint aligned with the GARI hypothesis: whether a generator-aligned interface preserves task information under held-out reversal without training on reversed samples. Because Caduceus is Mamba-based, we instantiate the AutoregressiveBlock in GARI-Net with Mamba2~\cite{Dao2024SSMDuality} in this experiment, so that the comparison remains within the same Mamba-based state-space backbone family rather than contrasting against an unrelated sequence block. In this diagnostic setting, however, no pretrained checkpoint is loaded for any model, so the comparison focuses on architecture-induced sequence-reversal robustness rather than representation transfer. GARI-Net uses the \texttt{gari-net\_small\_452k} setting, while Caduceus-PH and Caduceus-PS follow the PH and PS symmetry-handling variants from the Caduceus family; detailed layer counts and optimization hyperparameters are deferred to the appendix.

Table~\ref{tab:gb-data} summarizes the task statistics and the chance levels needed to interpret sequence-reversed evaluation. The suite contains both short fixed-length tasks, such as coding-vs-intergenic and human-vs-worm with length 200, and longer or variable-length regulatory tasks, such as enhancer, promoter, and OCR prediction. Binary tasks have a chance level of approximately 0.5, whereas the human regulatory task has three classes and a chance level of approximately 0.333.

\begin{table}[tbp]
  \centering
  \small
  \resizebox{\columnwidth}{!}{
    \begin{tabular}{lcccl}
      \toprule
      Task & Train & Classes & Max Len. & Length Regime \\
      \midrule
      Mouse enhancers & 1.2K & 2 & 1024 & Long, truncated \\
      Coding vs. intergenic & 100K & 2 & 200 & Short, fixed \\
      Human vs. worm & 100K & 2 & 200 & Short, fixed \\
      Human enhancers Cohn & 27.8K & 2 & 500 & Medium, fixed \\
      Human enhancer Ensembl & 154.8K & 2 & 512 & Variable \\
      Human regulatory & 289.1K & 3 & 512 & Variable, 3-class \\
      Human non-TATA promoters & 36.1K & 2 & 251 & Short, fixed \\
      Human OCR Ensembl & 174.8K & 2 & 512 & Variable \\
      \bottomrule
  \end{tabular}}
  \caption{GenomicBenchmarks tasks used in the sequence-equivariance enhancement diagnostic.}
  \label{tab:gb-data}
\end{table}

Table~\ref{tab:seq-equiv-gb-results} reports the sequence-level diagnostic results. The claim is that GARI-Net should preserve more task-relevant evidence under the sequence-reversal generator, not that it is a stronger genomic classifier in the ordinary identity-input setting. The control is the normal test accuracy, which checks whether all models learn the tasks to a comparable level before reversal. The primary endpoint is the full-suite flipped average. Because some tasks collapse to chance under reversal for all models, we also report an informative-task flipped average, but this parenthesized quantity is an auxiliary reading rather than a replacement for the all-task average. A task is treated as near-random when the mean flipped accuracy across the three models lies within the appropriate random baseline $\pm0.05$, i.e., $[0.45,0.55]$ for binary tasks and $[0.283,0.383]$ for the three-class task. This criterion removes human-vs-worm, human enhancer Ensembl, and human OCR Ensembl from the auxiliary informative-task flipped average, while retaining human regulatory because its three-class chance level is approximately 0.333. The supplementary material reports the sensitivity of this auxiliary average to the chance-band threshold. A result that would weaken the claim would be a higher flipped score explained only by higher normal accuracy, or an ordering that disappears on the full-suite flipped endpoint.

\begin{table}[tbp]
  \centering
  \caption{Sequence-reversal diagnostic on GenomicBenchmarks. Top-1 accuracy across five cross-validation seeds is reported as mean $\pm$ sample standard deviation, with best values within each evaluation setting bolded. Parenthesized values in the flipped Average row denote the informative-task flipped average computed on the same test split after excluding near-random sequence-reversed tasks, not a separate evaluation set.}
  \label{tab:seq-equiv-gb-results}
  \resizebox{\textwidth}{!}{
    \begin{tabular}{lcccccc}
      \toprule
      \multirow{2}{*}{Task} &
      \multicolumn{2}{c}{GARI-Net (452K)} &
      \multicolumn{2}{c}{Caduceus-PH (470K)} &
      \multicolumn{2}{c}{Caduceus-PS (470K)} \\
      \cmidrule(lr){2-3}\cmidrule(lr){4-5}\cmidrule(lr){6-7}
      & Normal & Flipped & Normal & Flipped & Normal & Flipped \\
      \midrule
      Mouse enhancers &
      \best{\score{0.777}{0.018}} & \best{\score{0.726}{0.012}} &
      \score{0.713}{0.006} & \score{0.717}{0.006} &
      \score{0.738}{0.019} & \score{0.689}{0.032} \\
      Coding vs. intergenic &
      \score{0.905}{0.002} & \best{\score{0.710}{0.007}} &
      \best{\score{0.905}{0.001}} & \score{0.686}{0.007} &
      \score{0.902}{0.001} & \score{0.709}{0.008} \\
      Human vs. worm &
      \score{0.960}{0.001} & \score{0.531}{0.007} &
      \best{\score{0.970}{0.001}} & \best{\score{0.551}{0.007}} &
      \score{0.968}{0.001} & \score{0.551}{0.006} \\
      Human enhancers Cohn &
      \score{0.720}{0.008} & \score{0.600}{0.027} &
      \best{\score{0.741}{0.003}} & \score{0.595}{0.020} &
      \score{0.738}{0.002} & \best{\score{0.604}{0.012}} \\
      Human enhancer Ensembl &
      \score{0.857}{0.003} & \score{0.487}{0.009} &
      \best{\score{0.882}{0.002}} & \score{0.490}{0.006} &
      \score{0.881}{0.003} & \best{\score{0.492}{0.004}} \\
      Human regulatory &
      \best{\score{0.875}{0.001}} & \best{\score{0.745}{0.009}} &
      \score{0.844}{0.030} & \score{0.700}{0.033} &
      \score{0.841}{0.035} & \score{0.695}{0.033} \\
      Human non-TATA promoters &
      \score{0.892}{0.020} & \best{\score{0.701}{0.049}} &
      \score{0.917}{0.007} & \score{0.634}{0.012} &
      \best{\score{0.938}{0.003}} & \score{0.610}{0.010} \\
      Human OCR Ensembl &
      \score{0.751}{0.019} & \score{0.521}{0.005} &
      \best{\score{0.819}{0.004}} & \score{0.526}{0.003} &
      \score{0.807}{0.003} & \best{\score{0.529}{0.004}} \\
      \midrule
      Average &
      0.842 & \best{0.627} (\best{0.696}) &
      0.849 & 0.612 (0.666) &
      \best{0.852} & 0.610 (0.661) \\
      \bottomrule
  \end{tabular}}
\end{table}

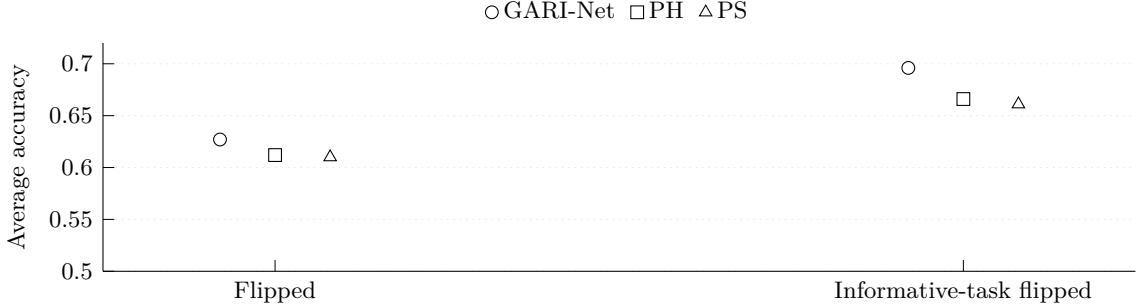
\begin{figure}[tbp]
  \centering
  \begin{tikzpicture}
    \begin{axis}[
        width=\columnwidth,
        height=4.6cm,
        xmin=-0.25,
        xmax=1.25,
        ymin=0.5,
        ymax=0.72,
        ytick={0.50,0.55,0.60,0.65,0.70},
        ylabel={Average accuracy},
        xtick={0,1},
        xticklabels={Flipped,{Informative-task flipped}},
        ymajorgrids=true,
        grid style={black!12,dotted},
        axis x line*=bottom,
        axis y line*=left,
        axis line style={black},
        tick style={black},
        legend style={at={(0.5,1.04)},anchor=south,legend columns=3,font=\footnotesize,draw=none,fill=none,/tikz/every even column/.append style={column sep=0.35em}},
        tick label style={font=\footnotesize},
        xticklabel style={align=center},
        label style={font=\footnotesize},
      ]
      \addplot+[only marks,black,mark=*,mark size=2.4pt,mark options={fill=white}] coordinates {(-0.08,0.627) (0.92,0.696)};
      \addplot+[only marks,black,mark=square*,mark size=2.4pt,mark options={fill=white}] coordinates {(0.00,0.612) (1.00,0.666)};
      \addplot+[only marks,black,mark=triangle*,mark size=2.7pt,mark options={fill=white}] coordinates {(0.08,0.610) (1.08,0.661)};
      \legend{GARI-Net,PH,PS}
    \end{axis}
  \end{tikzpicture}
  \caption{Dot-plot comparison of full-suite flipped average and informative-task flipped average for the sequence-reversal diagnostic. Points are slightly offset horizontally within each category only to avoid marker overlap. Informative-task averages exclude tasks whose mean flipped performance falls within the appropriate chance band $\pm0.05$.}
  \label{fig:gb-average}
\end{figure}

Before separating near-random and informative tasks, the normal-to-flipped averages already show that the models are comparably trained under the identity input but differ more clearly after sequence reversal: the full-suite flipped drop is $0.842-0.627=0.215$ for GARI-Net, compared with $0.849-0.612=0.237$ for Caduceus-PH and $0.852-0.610=0.242$ for Caduceus-PS.

The first observation is that sequence-reversed evaluation can expose a complete loss of discriminative signal even when normal accuracy is high. Human enhancer Ensembl and human OCR Ensembl fall close to binary chance for all three models, and human-vs-worm has a model-averaged flipped accuracy within the same random band. For example, human enhancer Ensembl has normal accuracy above 0.85 for all models but flipped accuracy around 0.49. These collapsed cases are important for the logic of this subsection: they show that standard task accuracy alone is insufficient to establish sequence-level equivariance, but they do not meaningfully distinguish the architectures once all models are near chance. We therefore exclude them from the informative-task flipped average.

On the remaining informative sequence-reversed tasks, GARI-Net provides stronger evidence of sequence-reversal robustness. It achieves the best full-suite flipped average, 0.627, and the best informative-task flipped average, 0.696, compared with 0.612/0.666 for Caduceus-PH and 0.610/0.661 for Caduceus-PS. As shown in Fig.~\ref{fig:gb-average}, removing near-random tasks raises all flipped averages but preserves the same ordering. At the task level, GARI-Net is best on mouse enhancers, coding-vs-intergenic, human regulatory, and human non-TATA promoters, while remaining competitive on human enhancers Cohn. Human regulatory is retained because its three-class chance level is approximately 0.333, so accuracies around 0.70--0.75 remain clearly informative rather than random.

The normal test set provides the intended control and reveals an explicit trade-off. Caduceus-PS obtains the best normal average, 0.852, followed by Caduceus-PH at 0.849 and GARI-Net at 0.842, so GARI-Net should not be described as the strongest genomic classifier under the identity input. The observed result is instead more specific: under comparable but slightly lower normal learning, GARI-Net has the best full-suite flipped average, 0.627, and the best auxiliary informative-task flipped average, 0.696. Thus the evidence is not that GARI-Net improves ordinary genomic classification, but that it better preserves reversal-transformed evidence once a task-relevant representation has been learned. This supports the sequence-side premise required by the subsequent sequence-to-vision equivariance analysis.

\subsubsection{\texorpdfstring{Cyclic-Generator Probes of $\mathrm{SO}(2)$ Rotation Robustness}{Cyclic-Generator Probes of SO(2) Rotation Robustness}}
This experiment operationalizes the practical reading of the equivariance analysis above. The theoretical development characterizes an equivariant limit: under ideal conditions, the induced vision model should commute with the group action. In a trained neural classifier, however, exact equality under all rotations is not expected, because the model is optimized from canonical images together with stochastic data augmentation rather than by imposing an explicit equivariance constraint for every group element. The relevant empirical question is therefore whether the proposed architecture improves \emph{soft equivariance}: by exposing the network to paired canonical and generator-transformed sequence views, GARI-Net makes group-consistent behavior easier to learn and thereby improves multi-angle generalization. We use $C_2$ and $C_4$ as controlled discretizations of the continuous rotation group to test whether the resulting rotation-generalization patterns follow the predicted group orbits. Since ImageNet classification uses rotation-invariant labels, output-level equivariance corresponds to prediction invariance under the trivial output representation; rotated Top-1 accuracy is therefore used as a task-level proxy for empirical equivariance and rotation generalization, not as a direct measurement of exact feature-space equivariance.

We compare Vision Mamba (VIM)~\cite{Zhu2024ViM} with two GARI-Net variants on ImageNet-1K. VIM is a natural baseline because it follows the same non-pyramidal ``PatchEmbed + Encoder'' design philosophy and uses a Mamba-based backbone, making the comparison closer than one against hierarchical vision architectures. It also provides a strong reference point among serialized vision models. The baseline is VIM-Small with about 27M parameters, which is the closest available VIM scale to the GARI-Net tiny configuration with about 30M parameters; the models are therefore closely matched in size. The two GARI-Net variants use the same implementation and differ only in the order of the cyclic generator supplied to the equivariant patch embedding: $\mathbb{Z}_2$ uses $n=2$, corresponding to the $C_2$ central-symmetry generator, whereas $\mathbb{Z}_4$ uses $n=4$, corresponding to the $C_4$ quarter-turn generator.

The VIM comparison should therefore be read as a mechanism-matched serialized-vision baseline, not as an exhaustive rotation-robustness leaderboard. The purpose is not to rank GARI-Net against every possible rotation-robust vision architecture, since such a comparison would require changing the feature hierarchy, inductive bias, parameter allocation, and often the training recipe. That would answer a different question about absolute robustness. Here the intended question is narrower and theory-driven: within the serialized vision regime, does adding generator-aligned patch-order structure produce the orbit-localized behavior predicted by the analysis? The ImageNet-1K experiment answers whether the interface scales to a demanding visual benchmark.

We include this clarification because many rotation-robust vision systems achieve performance by changing several factors simultaneously, including hierarchical feature construction, group-specific operators, augmented training recipes, or task-specific invariance assumptions. Such comparisons are important for practical deployment but are not the cleanest test of the GARI hypothesis. The present ImageNet-1K endpoint instead asks whether, within a closely matched serialized-vision setting, changing the exposed cyclic generator produces the predicted orbit-localized behavior.

Accordingly, this ImageNet study is not intended to isolate every architectural factor in GARI-Net. Its role is to test whether, against a strong serialized-vision baseline, changing the cyclic generator produces the orbit-structured rotation behavior predicted by the analysis. The corresponding same-architecture generator-coverage control is reported in the following MNIST experiment, where the backbone is fixed and the number of GARI-Net generators is varied, including the zero-generator setting.

All models are trained from scratch using three random seeds, $0$, $42$, and $625$, under the same ImageNet-1K recipe. Each run uses $224\times224$ inputs, a total batch size of 1024, AdamW optimization for 300 epochs, cosine learning-rate decay, Mixup/CutMix, RandomErasing, RandAugment, and EMA weights for validation. The only differences across the runs are the model choice and, for GARI-Net, the generator order $n$. We use RandAugment magnitude $m=20$, whose rotation operation has an effective maximum magnitude of approximately $\pm60^\circ$. This relatively strong augmentation is intentional: it reduces overfitting caused by the group-structured patch embedding and gives VIM a nontrivial level of rotation exposure, making it a meaningful baseline rather than a model that fails under rotation by construction. At test time, ImageNet-1K validation images are rotated by angles from $0^\circ$ to $345^\circ$ in $15^\circ$ increments using the same deterministic rotation pipeline for all models, with fixed interpolation, padding/fill, resizing/cropping, and normalization settings. Top-1 accuracy is then summarized across the three seeds. Because each 300-epoch ImageNet-1K run is computationally expensive, we use a paired three-seed design rather than a five-seed design. To keep the statistical interpretation conservative, all reported improvements over VIM are computed seedwise before averaging, and the conclusions rely on pre-specified orbit regions and angle-resolved localization rather than on post-hoc significance claims from a small number of large-scale runs.

The $\pm60^\circ$ augmentation range defines nine augmentation-supported evaluation angles, $\{0^\circ,15^\circ,30^\circ,45^\circ,60^\circ,300^\circ,315^\circ,330^\circ,345^\circ\}$. The remaining fifteen angles from $75^\circ$ to $285^\circ$ form the pure-generalization interval. Under a trivial group action, VIM has no explicit orbit coverage outside the observed augmentation range. Under the $C_2$ generator, the augmentation-supported angles are mapped by $180^\circ$ rotation to the interval $120^\circ$--$240^\circ$, covering nine of the fifteen pure-generalization angles. Under the $C_4$ generator, quarter-turn rotations map the same augmentation-supported set to all twenty-four evaluated angles, so the pure-generalization interval is fully covered by the induced orbit structure. This protocol directly tests whether the empirical gains follow the algebraic coverage implied by the chosen generator.

The primary endpoint for this experiment is therefore the pure-generalization interval, not the all-rotation average. The all-rotation average is retained as a diluted control because it mixes held-out angles with augmentation-supported angles. A result that would weaken the mechanism claim would be a gain concentrated only inside the augmentation interval, or an orbit-localization pattern that does not follow the $C_2/C_4$ coverage.

\begin{table}[tbp]
  \centering
  \caption{Primary pure-generalization and orbit-specific cyclic-generator results on ImageNet-1K. Top-1 accuracy is reported as mean $\pm$ sample standard deviation over three seed-level region averages. $\Delta$ denotes the mean paired improvement over VIM, computed with the same seed and then averaged over the three seeds, in percentage points. The primary readouts are the pure-generalization interval and the generator-orbit regions, which exclude or isolate angles outside the augmentation-supported range. The augmentation interval and the all-rotation average are diluted controls, not primary endpoints.}
  \label{tab:imagenet-so2-results}
  \resizebox{\textwidth}{!}{%
    \begin{tabular}{lcccccc}
      \toprule
      Evaluation region & Angle set & VIM & $C_2$ & $C_2$ $\Delta$ & $C_4$ & $C_4$ $\Delta$ \\
      \midrule
      Pure generalization (primary) & $75^\circ$--$285^\circ$ (15 angles) & \score{52.85}{0.85} & \score{53.85}{0.23} & +1.00 & \best{\score{54.19}{1.06}} & \best{+1.34} \\
      $C_2$ orbit coverage & $120^\circ$--$240^\circ$ (9 angles) & \score{52.29}{0.77} & \best{\score{54.78}{0.32}} & \best{+2.49} & \score{52.52}{0.91} & +0.24 \\
      $C_4$-only orbit coverage & $75^\circ/90^\circ/105^\circ/255^\circ/270^\circ/285^\circ$ & \score{53.70}{0.97} & \score{52.46}{0.37} & -1.24 & \best{\score{56.70}{1.32}} & \best{+3.00} \\
      Augmentation interval (control) & $\pm60^\circ$ (9 angles) & \best{\score{71.31}{0.21}} & \score{69.79}{0.47} & -1.52 & \score{70.89}{0.64} & -0.42 \\
      All rotations (diluted control) & $0^\circ$--$345^\circ$ (24 angles) & \score{59.77}{0.61} & \score{59.83}{0.32} & +0.05 & \best{\score{60.45}{0.90}} & \best{+0.68} \\
      \bottomrule
  \end{tabular}}
\end{table}

Complete angle-wise seed values for the ImageNet-1K rotation study are reported in the supplementary material. VIM's stronger score on the augmentation-supported interval is an expected trade-off under the strong RandAugment rotation exposure and is treated as a control, not as a failure of the held-out generator-localized endpoint.

\begin{figure}[tbp]
  \centering
  \begin{minipage}{0.48\textwidth}
    \centering
    \includegraphics[width=\linewidth]{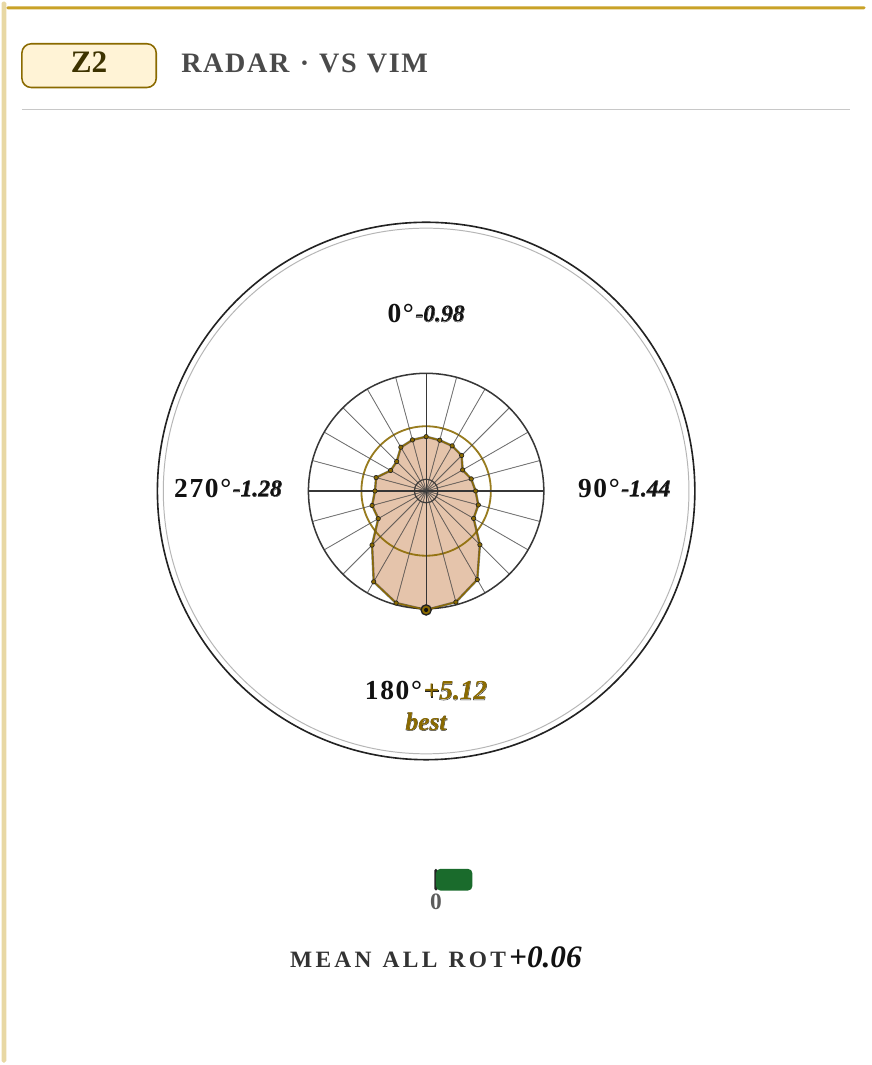}
    \par\smallskip\small (a) $\mathbb{Z}_2$ versus VIM.
  \end{minipage}
  \hfill
  \begin{minipage}{0.48\textwidth}
    \centering
    \includegraphics[width=\linewidth]{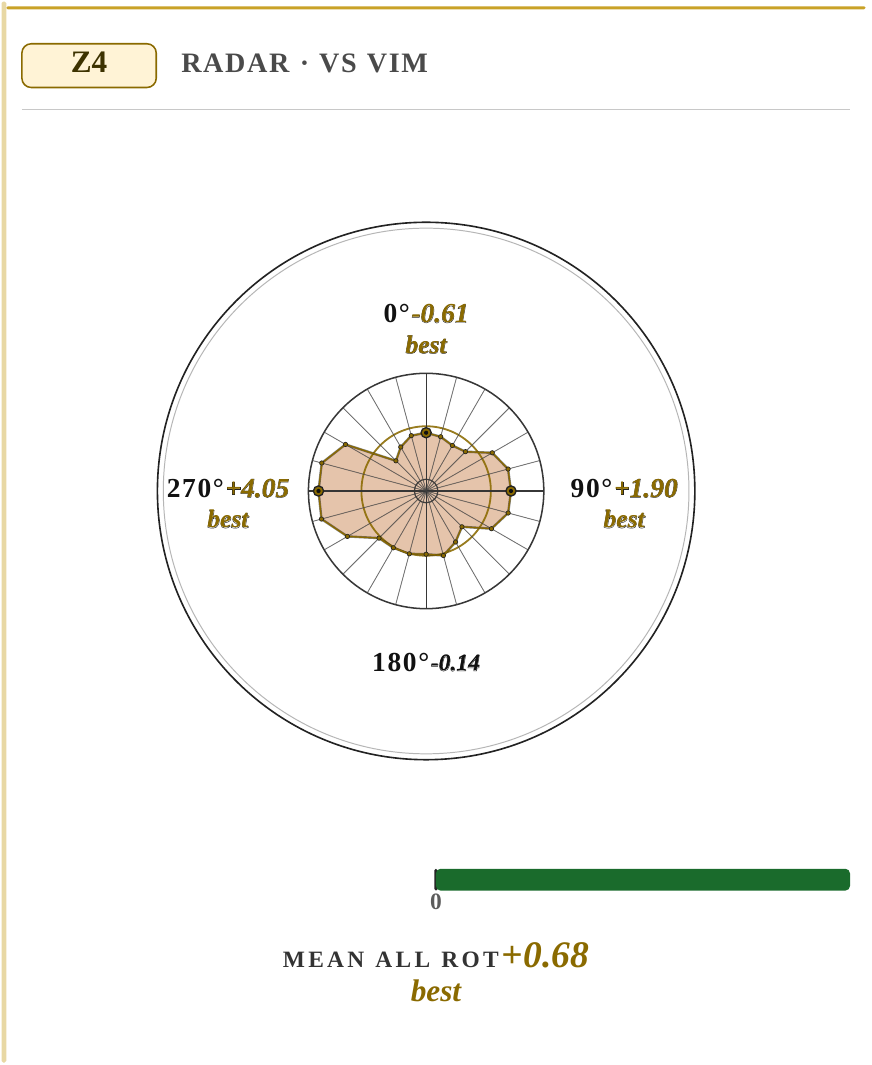}
    \par\smallskip\small (b) $\mathbb{Z}_4$ versus VIM.
  \end{minipage}
  \caption{Pure-generalization and orbit-specific radar summaries for the ImageNet-1K cyclic-generator rotation experiment. Each panel reports the Top-1 accuracy difference relative to VIM at the four cardinal axes; the all-rotation mean is shown only as a diluted control. The $\mathbb{Z}_2$ model concentrates its gain around the $180^\circ$ orbit direction, whereas the $\mathbb{Z}_4$ model distributes gains across the quarter-turn directions. The reader should prioritize these held-out and generator-localized gains rather than the diluted all-rotation average.}
  \label{fig:imagenet-so2-radar}
\end{figure}

To provide a direct representation-level diagnostic beyond rotation accuracy, we also report a post-hoc Direct Equivariance Error (DEE) on frozen ImageNet-1K checkpoints. Let $h_\ell(x)$ denote the pre-pooling patch-token feature at layer $\ell$ after removing any classification token, let $\mathcal{N}$ be channel-wise layer normalization, and let $\rho_\alpha$ be the scan-induced token permutation corresponding to the image rotation $R_\alpha$. Operationally, the representation action $\tau_g$ for this DEE is this known token permutation; no fitted representation action is learned. For each evaluated rotation, we compute
\begin{equation}
  \label{eq:dee}
  \mathrm{DEE}^{(\ell)}(R_\alpha)
  =
  \mathbb{E}_{x}
  \frac{
    \left\lVert
    \mathcal{N}\!\left(h_\ell(R_\alpha x)\right)
    -
    \rho_\alpha\mathcal{N}\!\left(h_\ell(x)\right)
    \right\rVert_F
  }{
    \frac{1}{2}\left(
      \left\lVert\mathcal{N}\!\left(h_\ell(R_\alpha x)\right)\right\rVert_F
      +
      \left\lVert\mathcal{N}\!\left(h_\ell(x)\right)\right\rVert_F
    \right)+\varepsilon
  }.
\end{equation}
Lower DEE indicates a more group-consistent internal representation. We report $\mathrm{DEE}_{180}=\mathrm{DEE}(R_{180})$ and
\begin{equation}
  \label{eq:dee-c4}
  \mathrm{DEE}_{C_4}
  =
  \frac{1}{3}\left[
    \mathrm{DEE}(R_{90})+
    \mathrm{DEE}(R_{180})+
    \mathrm{DEE}(R_{270})
  \right].
\end{equation}
This metric is not used for training; it is only a frozen-model representation-level consistency diagnostic. It can align with task-level rotation robustness, but it should not be interpreted as fully explaining the accuracy gain.

\begin{table}[tbp]
  \centering
  \caption{Post-hoc Direct Equivariance Error (DEE) on ImageNet-1K frozen checkpoints. Early denotes the output of the first encoder block, and final denotes the final pre-pooling patch-token representation. Lower DEE is better. Top-1 columns repeat the primary pure-generalization endpoint and the diluted all-rotation control from \Cref{tab:imagenet-so2-results} to show how representation-level consistency aligns with held-out rotation generalization.}
  \label{tab:imagenet-so2-dee}
  \resizebox{\textwidth}{!}{%
    \begin{tabular}{llcccccc}
      \toprule
      Model & Generator & Early $\mathrm{DEE}_{180}$ & Early $\mathrm{DEE}_{C_4}$ & Final $\mathrm{DEE}_{180}$ & Final $\mathrm{DEE}_{C_4}$ & Pure-gen. Top-1 & All-rot. Top-1 \\
      \midrule
      VIM & -- & \best{0.715} & \best{0.776} & 0.981 & 0.983 & 52.85 & 59.77 \\
      GARI-Net-$C_2$ & $C_2$ & 0.912 & 0.990 & \best{0.771} & 0.835 & 53.85 & 59.83 \\
      GARI-Net-$C_4$ & $C_4$ & 0.802 & 0.900 & 0.808 & \best{0.831} & \best{54.19} & \best{60.45} \\
      \bottomrule
  \end{tabular}}
\end{table}

The primary message of Table~\ref{tab:imagenet-so2-results} is the pure-generalization and orbit-specific gain, not the all-rotation average. The pure-generalization interval excludes the $\pm60^\circ$ angles directly supported by RandAugment; it is therefore the relevant endpoint for testing whether generator-aligned structure extrapolates beyond the training distribution. On this primary interval, $C_4$ improves over VIM by $+1.34$ percentage points and $C_2$ improves by $+1.00$ percentage point. The all-rotation mean is reported only as a diluted control: it mixes these held-out angles with augmentation-supported angles where VIM remains the strongest model, so it necessarily suppresses the extrapolation signal that is the actual endpoint of this experiment.

This control is useful because it separates in-distribution rotation adaptation from extrapolation. The fact that VIM remains slightly stronger in the augmentation interval is an expected trade-off rather than a weakness to hide: ordinary augmentation gives the baseline direct exposure to those angles, while GARI-Net allocates capacity to generator-structured extrapolation. The relevant differences therefore emerge mainly outside the $\pm60^\circ$ range, where the generator supplies an explicit relation between the canonical patch ordering and its transformed ordering. Improving this out-of-augmentation regime is a stronger mechanism claim than improving the global average, because the model must transfer a learned group-action rule to angles not directly optimized during training.

The relatively larger standard deviations in the pure-generalization and generator-specific unseen regions should be interpreted in this extrapolation setting. These regions consist of angles outside the rotation range directly supported by training augmentation, so the reported accuracies measure generalization to unseen rotations rather than interpolation among angles observed during training. Small seed-dependent differences in the learned canonical classifier and in how strongly the generator-induced patch-order correspondence is internalized can therefore be amplified at these unseen angles, producing larger across-seed variability than in the augmentation interval. In addition, the region averages combine rotation angles with different intrinsic difficulty, so the standard deviation should not be read as only optimizer-induced seed noise. Increasing the number of seeds would reduce the seed-estimation component, but it would not remove the angle-distribution component of this extrapolation protocol. For this reason, the primary evidence is not the variance of any single unseen-angle average alone, but the conjunction of three safeguards: paired seedwise deltas, pre-specified orbit-coverage regions, and angle-resolved gain localization matching the predicted group structure.

This contrast also clarifies the limitation of ordinary rotation augmentation for serialized vision models. Rotation augmentation presents different rotated views as independent training samples, but it does not explicitly provide the correspondence between the original patch sequence and the rotated patch sequence, nor does it specify how the rotation group should act consistently on the image grid and feature space. Therefore, a model such as VIM may adapt to the sequence distributions induced by observed rotation angles without learning a coherent group-action rule that extrapolates to unseen or compositional rotations. GARI-Net addresses this limitation by injecting group structure at the patch-embedding level and by applying the corresponding generator reindexing within the sequential processing path, so that the two streams process complementary geometric orderings of the same image.

For the $C_2$ generator, the observed advantage is concentrated exactly where the orbit analysis predicts. In the $120^\circ$--$240^\circ$ interval, $C_2$ improves over VIM by $+2.49$ percentage points on average, and Fig.~\ref{fig:imagenet-so2-radar}(a) shows a peak gain of $+5.12$ percentage points at $180^\circ$. However, the six angles $75^\circ$, $90^\circ$, $105^\circ$, $255^\circ$, $270^\circ$, and $285^\circ$ are not covered by the $C_2$ orbit of the augmentation-supported angles; on this blind region, $C_2$ is lower than VIM by $1.24$ percentage points on average. The $C_2$ result therefore supports a directional-robustness interpretation rather than a claim of uniform planar rotation robustness.

The $C_4$ generator yields broader and more balanced held-out gains. Since the $C_4$ orbit of the augmentation-supported angles covers all twenty-four evaluation angles, $C_4$ has no orbit blind region under this protocol. Empirically, it improves by $+3.00$ percentage points on the $C_4$-only region and by $+1.34$ percentage points over the full pure-generalization interval. The all-rotation improvement of $+0.68$ percentage points is therefore a diluted control, not the main conclusion, because it adds back angles where the baseline already receives direct augmentation support. $C_4$'s gain in the $C_2$-covered interval is smaller than that of $C_2$, suggesting a trade-off between concentrating capacity on one dominant generator direction and distributing the geometric bias across a larger orbit.

The cardinal-axis results provide an additional implementation-level check. For $C_4$, the largest gain is at $270^\circ$ rather than $90^\circ$, with improvements of $+4.05$ and $+1.90$ percentage points, respectively. This asymmetry is consistent with the actual generator direction used by \textsf{GeneratorReindex}: with \texttt{inverse=false}, the backward-chain token permutation maps $(\mathrm{row},\mathrm{col})$ to $(\mathrm{col},N-1-\mathrm{row})$, corresponding to a clockwise $90^\circ$ rotation, i.e., $270^\circ$ under the counterclockwise evaluation convention. Thus, the strongest empirical gain aligns with the direction explicitly processed by the generator-ordered stream, providing a useful sanity check that the measured robustness is tied to the intended sequence-level group action.

The DEE diagnostic in \Cref{tab:imagenet-so2-dee} gives a complementary view of where this group consistency appears inside the network. At the early layer, VIM has lower DEE than either GARI-Net variant. This does not contradict the proposed mechanism: early patch-near features are strongly affected by the expressivity--equivariance trade-off of the patch embedding. As shown by the cyclic kernel analysis in the methodology and detailed in the supplementary material, exact equivariance of a convolutional patch embedding requires the kernel to be invariant under the target group action, and increasing the rotation order further restricts admissible kernels toward increasingly isotropic, radial structures. A low early-layer DEE may therefore reflect local isotropic or weakly directional low-level statistics rather than a semantically useful group-consistent representation. GARI-Net deliberately avoids forcing the raw PatchEmbed to be exactly equivariant so that the early visual interface can retain kernel diversity and discriminative directional features.

The decisive pattern appears in the final pre-pooling representation. VIM has a high final $\mathrm{DEE}_{C_4}$ of $0.983$, indicating that its early apparent equivariance does not propagate into the final semantic representation. By contrast, GARI-Net-$C_2$ and GARI-Net-$C_4$ reduce final $\mathrm{DEE}_{C_4}$ to $0.835$ and $0.831$, respectively, matching the ordering of the pure-generalization Top-1 scores. The group specificity also follows the generator design: GARI-Net-$C_2$ has the lowest final $\mathrm{DEE}_{180}$, $0.771$, consistent with its $C_2$ generator, whereas GARI-Net-$C_4$ has the lowest final $\mathrm{DEE}_{C_4}$, $0.831$, consistent with broader coverage of the $90^\circ/180^\circ/270^\circ$ orbit. Thus, DEE is consistent with the same mechanism suggested by the accuracy results: GARI-Net need not make the raw patch embedding strictly equivariant; instead, it organizes equivariance progressively through the generator-aligned representation interface. This representation-level consistency aligns with, but does not fully explain, the stronger held-out rotation generalization.

Overall, the experiment supports the intended interpretation of GARI-Net as a generator-informed geometric learning architecture, while not by itself ruling out all effects of the two-stream architecture or the small remaining parameter-count mismatch. This remaining alternative explanation is the target of the following controlled generator-count study, which fixes the backbone and includes a zero-generator control. The model does not hard-code exact continuous $\mathrm{SO}(2)$ equivariance; rather, finite cyclic generators provide structured geometric views that encourage soft equivariance under the corresponding group action. The agreement between orbit coverage and empirical gains across $C_2$ and $C_4$ suggests that improving this group-consistent behavior is a plausible mechanism by which GARI-Net improves task-level empirical equivariance and planar rotation generalization.

\subsubsection{\texorpdfstring{Generator-Count Effects on $\mathrm{O}(2)$ Equivariance}{Generator-Count Effects on O(2) Equivariance}}
This experiment provides the same-architecture no-generator control that complements the ImageNet comparison above. It asks whether exposing generators of the planar symmetry group to GARI-Net improves generalization to a sampled set of $\mathrm{O}(2)$ transformations when the backbone, parameter count, and training recipe are controlled. We use MNIST because the small image size makes it practical to evaluate a fixed transformation orbit for every trained model. Importantly, GARI-Net is not a hard-constrained group-equivariant network. We therefore use accuracy along this finite transformation set as a behavioral proxy for learned equivariance: a generator-aware model should degrade less, on average and in the worst case, when the input is transformed by rotations and a held-out reflection that were not all seen during training. As in the ImageNet rotation study, the clean score is a control rather than the main endpoint; the primary question is whether the architecture improves out-of-augmentation transformations.

We compare five GARI-Net variants and a parameter-matched ViT baseline. All GARI-Net models use native $28\times28$ MNIST images, a $4\times4$ patch stem, embedding dimension $104$, depth $8$, and a GRU sequence core. $C_0$ is the internal no-generator control: it uses the same GARI-Net backbone but a single ordinary patch-embedding stream, with no group action. $C_2$ and $C_4$ use a two-stream cyclic backend with an identity stream and one rotation-generator stream; $n=2$ applies the $180^\circ$ central-symmetry generator, while $n=4$ applies a $90^\circ$ rotation generator. $D_2$ and $D_4$ use the corresponding three-stream backend, adding a reflected stream. After each generated view is formed, its patch sequence is re-indexed back to the canonical raster order, and the streams interact through the same GARI-Net encoder and cross-stream attention before classification. Thus the variants expose different group structure while keeping the downstream architecture as fixed as possible.

The baselines serve different roles. $C_0$ controls for the GARI-Net backbone, optimizer, patch stem, and sequence core, leaving the presence of generator streams as the main intervention. ViT is an external isotropic embedding-plus-encoder reference rather than a deliberately weak comparator; it uses the same patch size and is parameter-matched to the same $\approx1.02$M scale. Since the GARI-Net sequence core in this experiment is a GRU rather than a stronger self-attention block, improvements over ViT should not be attributed to a more expressive backbone. Moreover, the GARI-Net streams share encoder weights: $C_2$/$C_4$ add only $18$ parameters relative to $C_0$, and $D_2$/$D_4$ add about $0.5\%$ more parameters. The comparison therefore primarily changes whether generator-structured streams are exposed to the model, not model capacity.

The controlled factors are separated as follows. Generator order is tested by $C_2$ versus $C_4$: both use one rotation-generator stream and therefore have matched parameters and FLOPs, so their difference isolates cyclic order and orbit coverage. Stream count is tested by comparing cyclic $C$-variants with dihedral $D$-variants: this exposes an additional reflection generator but is not FLOP-matched. The no-generator control is $C_0$, which keeps the GARI-Net backbone, patch stem, optimizer, and sequence core while removing generator streams. The primary endpoint is transformed-orbit robustness, not clean accuracy. A result that would weaken the claim would be a gain explained only by extra streams or clean accuracy rather than by generator order or held-out transformations.

\begin{table}[htbp]
  \centering
  \small
  \begin{tabular}{llrrrr}
    \toprule
    Model & Backend & Streams & Params & FLOPs (M) & MACs (M) \\
    \midrule
    $C_0$ & \texttt{SingleStream} & 1 & 1,019,290 & 97.7 & 48.9 \\
    $C_2$ & \texttt{standard} & 2 & 1,019,308 & 191.1 & 95.5 \\
    $C_4$ & \texttt{standard} & 2 & 1,019,308 & 191.1 & 95.5 \\
    $D_2$ & \texttt{ThreeStreams} & 3 & 1,024,716 & 284.9 & 142.5 \\
    $D_4$ & \texttt{ThreeStreams} & 3 & 1,024,716 & 284.9 & 142.5 \\
    ViT & -- & -- & 1,019,026 & 65.6 & 32.8 \\
    \bottomrule
  \end{tabular}
  \caption{Parameter and compute profile for the MNIST $\mathrm{O}(2)$ controlled experiment. Parameter counts are read from \texttt{logs/training/*.out}; FLOPs are measured by \texttt{scripts/model\_flops.py} with PyTorch's \texttt{FlopCounterMode} for one batch-size-one $1\times28\times28$ image, and MACs are reported as FLOPs$/2$.}
  \label{tab:mnist-o2-compute}
\end{table}

Table~\ref{tab:mnist-o2-compute} makes the capacity and compute trade-off explicit. The parameter counts are intentionally matched: cyclic generator variants add only 18 parameters over $C_0$, and the dihedral variants add about $5.4$K parameters, roughly $0.5\%$ of the model size. Thus, the robustness differences below should not be interpreted as ordinary capacity scaling. FLOPs, however, are not matched across different stream counts. Because each exposed stream executes the shared encoder on a different generated view, inference compute grows approximately with the number of exposed generator streams: $C_0$ requires $97.7$M FLOPs, $C_2$/$C_4$ require $191.1$M, and $D_2$/$D_4$ require $284.9$M. A key observation is that compute is controlled by the number of generators, not by the order of a cyclic generator. $C_2$ and $C_4$ have identical parameter counts and FLOPs because both use one rotation-generator stream; the difference between them is the order of the rotation generator and hence the induced orbit coverage. Similarly, $D_2$ and $D_4$ have identical parameter counts and FLOPs because both add the same reflection stream. The controlled claim in this experiment is therefore parameter-matched rather than FLOP-matched: GARI-Net trades additional forward compute for improved sampled-$\mathrm{O}(2)$ generalization when more independent generator streams are exposed. FLOPs also should not be read as wall-clock latency, since the GRU core processes the 49-token patch sequence sequentially whereas the ViT reference can parallelize attention over the sequence dimension. The MNIST result therefore does not claim compute superiority; rather, it shows that under matched parameter capacity and a comparatively simple sequential core, exposing generator-aligned streams improves the relevant orbit-robustness endpoint.

All models are trained for 10 epochs on MNIST with Adam, learning rate $10^{-3}$, batch size 128, no weight decay, and no dropout/drop-path. Training augmentation uses random rotations within $\pm60^\circ$ and horizontal flips with probability $0.5$. We train three seeds, $42$, $1$, and $2$. At evaluation time, each checkpoint is tested on the clean MNIST test set transformed by twelve rotations, $0^\circ,30^\circ,\ldots,330^\circ$, together with one vertical flip. The rotation test therefore includes angles far outside the training range, and the vertical flip deliberately uses a different reflection axis from the horizontal flip seen during training. These held-out transformations probe generalization over this sampled set of $\mathrm{O}(2)$ transformations rather than direct memorization of the training augmentation. Rotations are evaluated with bilinear interpolation and zero fill, and the evaluation transform is deterministic and independent of the training seed. We report Top-1 accuracy averaged over seeds; the orbit-average standard deviation in Table~\ref{tab:mnist-o2-summary} is computed by first averaging each seed over the thirteen evaluated transformations and then taking the sample standard deviation across seeds.

\begin{table}[tbp]
  \centering
  \caption{Primary held-out-orbit and worst-case generalization on a sampled set of $\mathrm{O}(2)$ transformations for MNIST. Clean denotes the $0^\circ$ test accuracy and is reported only as a control. Orbit average is computed over twelve rotations and one vertical flip, many of which are outside the augmentation distribution, and worst case is the minimum accuracy over the same thirteen transformations. $\Delta$ values are orbit-average improvements in percentage points and are interpreted as held-out-orbit generalization gains rather than clean-accuracy gains.}
  \label{tab:mnist-o2-summary}
  \resizebox{\textwidth}{!}{%
    \begin{tabular}{lcccccccc}
      \toprule
      Model & Exposed group & Streams & Params & Clean & Orbit avg. & Worst case & $\Delta$ vs $C_0$ & $\Delta$ vs ViT \\
      \midrule
      $C_0$ & None & 1 & 1.019M & 93.84 & \score{62.16}{0.83} & 38.71 & -- & +0.18 \\
      $C_2$ & $C_2$ & 2 & 1.019M & 93.91 & \score{62.85}{0.54} & 38.87 & +0.69 & +0.87 \\
      $C_4$ & $C_4$ & 2 & 1.019M & \best{95.34} & \best{\score{64.77}{0.52}} & \best{41.76} & \best{+2.61} & \best{+2.79} \\
      $D_2$ & $D_2$ & 3 & 1.025M & 93.20 & \score{63.24}{1.40} & 40.44 & +1.08 & +1.26 \\
      $D_4$ & $D_4$ & 3 & 1.025M & 94.22 & \score{63.46}{0.20} & 39.90 & +1.30 & +1.48 \\
      ViT & None & -- & 1.019M & 94.03 & \score{61.98}{0.87} & 37.57 & -0.18 & -- \\
      \bottomrule
  \end{tabular}}
\end{table}

The primary message of Table~\ref{tab:mnist-o2-summary} is the held-out-orbit average and worst-case transformed accuracy, not the clean control. All generator-aware GARI-Net variants improve the orbit average over the same-architecture $C_0$ control and over the parameter-matched ViT reference, despite having nearly the same parameter count. As shown in Table~\ref{tab:mnist-o2-compute}, this is not a FLOP-matched comparison across stream counts: the multi-stream variants spend additional inference compute to expose generator-structured views. However, the order comparison within a fixed generator count is compute-matched. $C_2$ and $C_4$ have the same parameter count and FLOPs, so the stronger $C_4$ result isolates the effect of higher cyclic order and broader orbit coverage rather than additional compute. The strongest variant is $C_4$: it obtains the best orbit average and the best worst-case accuracy, while also maintaining the best clean score. Its orbit average is $+2.61$ percentage points above $C_0$ and $+2.79$ percentage points above ViT, while its worst-case accuracy is higher by about $+3.05$ and $+4.19$ percentage points, respectively. Thus the main generalization gain is not obtained by ordinary capacity scaling, and it is not merely a clean-accuracy effect; it appears on transformed inputs that are outside, or mismatched with, the augmentation distribution. Additional independent generator streams do come with the expected multi-stream compute overhead.

To further examine how the exposed group structure affects the equivariance of the learned representation, we additionally report DEE in the MNIST controlled group-coverage experiment. Following the ImageNet-1K diagnostic in \Cref{eq:dee}, DEE is computed post hoc on frozen checkpoints and is not used as a training objective. Operationally, $\tau_g$ is the known token reindexing induced by each evaluated rotation or reflection; no fitted representation action is used. For each model, we measure the average equivariance error on the final pre-pooling token representation under four finite transformation averages:
\begin{equation}
  \label{eq:mnist-o2-dee-groups}
  \begin{aligned}
    \mathrm{DEE}_{C_2} &= \mathrm{DEE}(R_{180}),\\
    \mathrm{DEE}_{C_4} &= \frac{1}{3}\sum_{\alpha\in\{90,180,270\}}\mathrm{DEE}(R_\alpha),\\
    \mathrm{DEE}_{D_2} &= \frac{1}{3}\left[\mathrm{DEE}(R_{180})+\mathrm{DEE}(H)+\mathrm{DEE}(V)\right],\\
    \mathrm{DEE}_{D_4} &= \frac{1}{7}\sum_{g\in D_4\setminus\{e\}}\mathrm{DEE}(g),
  \end{aligned}
\end{equation}
where $H$ and $V$ denote horizontal and vertical flips, respectively. The GARI-Net variants have approximately matched capacity and mainly differ in the group structure exposed to the architecture: none, $C_2$, $C_4$, $D_2$, or $D_4$.

Before interpreting the DEE table, we distinguish numerical consistency from discriminative equivariance. Low DEE alone is not sufficient evidence for the desired behavior: a representation can be smooth or partially invariant while still failing to classify transformed inputs. The positive evidence sought here is lower relevant DEE together with improved transformed-orbit accuracy and worst-case accuracy. This distinction is important because the $C_0$ no-generator control can have low DEE on some group averages without being the most robust classifier.

The complete MNIST final-representation DEE breakdown is reported in the supplementary material; the main text uses it as a secondary diagnostic for the task-level orbit results.

The supplementary DEE results support the same joint interpretation as the task metrics, but also clarify the scope of the diagnostic. $C_4$ achieves the lowest error on the most relevant rotation-coverage metric, $\mathrm{DEE}_{C_4}=1.0056$, and it also obtains the best orbit-average accuracy, $64.77\%$, and worst-case accuracy, $41.76\%$. This indicates that explicitly exposing the $C_4$ generator improves the consistency of the final representation over the $90^\circ/180^\circ/270^\circ$ rotation orbit, and that this representation-level improvement is accompanied by stronger orbit-level generalization. In contrast, $C_2$ only exposes the $C_2$ generator and performs worse than $C_4$ in both $\mathrm{DEE}_{C_4}$ and the orbit-level task metrics, suggesting that broader cyclic generator coverage is beneficial for rotation-orbit robustness.

At the same time, the no-generator $C_0$ control even obtains the lowest DEE on several group averages, including $\mathrm{DEE}_{C_2}$, $\mathrm{DEE}_{D_2}$, and $\mathrm{DEE}_{D_4}$. This highlights an important distinction between numerical representation consistency and discriminative equivariance. DEE measures the feature-space residual under a group action, but it does not by itself guarantee that the representation is non-degenerate or sufficiently discriminative. A model may learn smoother, more invariant, or less direction-sensitive features that reduce the numerical discrepancy between transformed inputs, without achieving strong classification performance over the transformation orbit. This is reflected by $C_0$: although it attains the lowest DEE on several averages, its orbit-average accuracy, $62.16\%$, and worst-case accuracy, $38.71\%$, remain clearly below those of $C_4$. Therefore, we interpret DEE as a direct representation diagnostic rather than a replacement for the task metric. Lower DEE is most meaningful when it is accompanied by improved orbit-level accuracy. Under this joint interpretation, $C_4$ provides the strongest positive evidence in this experiment: it achieves the lowest $C_4$-rotation equivariance error while also obtaining the best orbit-average and worst-case accuracy. These results support the claim that exposing a more complete cyclic generator structure improves both the group consistency of the final representation and its robustness to transformation-orbit shifts.

The location of the gains is also consistent with the generator interpretation and highlights the deliberate train/test mismatch. During training, the model only sees rotations within $\pm60^\circ$ together with horizontal flips, whereas evaluation includes large rotations from $90^\circ$ through $270^\circ$ and a held-out vertical reflection. These transformations are outside, or aligned along a different reflection axis from, the augmentation distribution, so they probe generalization along the sampled $\mathrm{O}(2)$ orbit rather than recall of augmented training views. This is precisely the regime in which learned equivariance should help. The largest $C_4$ improvements occur in this mismatched sector, especially around far rotations such as $180^\circ$ and $270^\circ$, and the vertical-flip result also improves over both no-generator baselines. Paired tests over the thirteen evaluated transformations give the same qualitative conclusion for the internal no-generator control: all generator-aware variants improve over $C_0$ at $p<0.05$, with $C_4$ and $D_4$ showing the strongest paired evidence. Relative to ViT, $C_4$ is the clearest case, improving on all thirteen transformations; the smaller $C_2$-vs-ViT gain is positive in mean but is not statistically significant under the same paired tests.

The dihedral variants provide a useful qualification about adding a second generator. Relative to the no-generator $C_0$ control, $D_2$ and $D_4$ improve both the orbit average and the held-out vertical reflection, whereas the $C_2$ cyclic model is essentially unchanged on the flip. This suggests that the reflection stream can improve generalization in the added-generator direction relative to a no-generator GARI-Net control. This should not be read as a claim that two generators uniformly dominate one generator: $C_4$ remains the strongest overall variant, and fixed capacity must be allocated across more induced streams. Rather, the result indicates that exposing an additional generator can target the corresponding out-of-augmentation direction while preserving robustness on the sampled $\mathrm{O}(2)$ orbit under matched parameter scale.

\begin{figure}[tbp]
  \centering
  \begingroup
  \setlength{\tabcolsep}{1.5pt}
  \renewcommand{\arraystretch}{0.5}
  \resizebox{0.58\textwidth}{!}{%
    \begin{tabular}{cc}
      \includegraphics[width=0.40\textwidth]{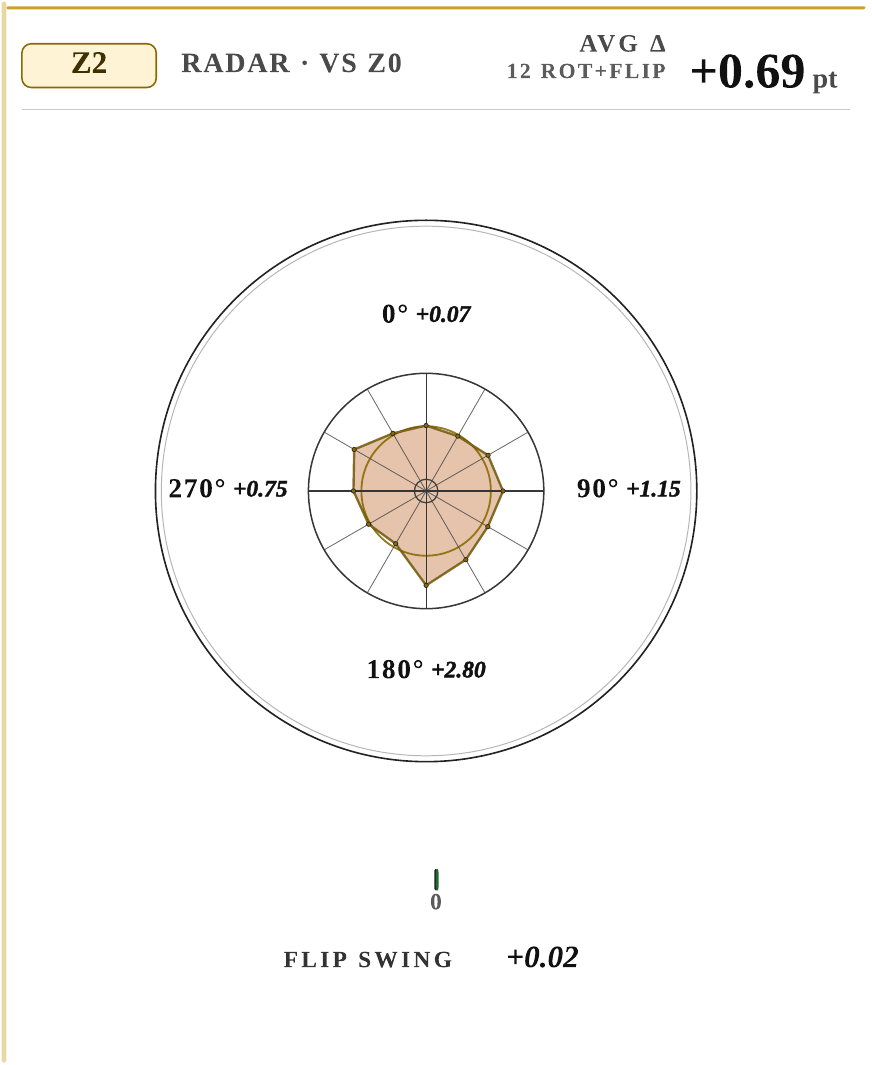} &
      \includegraphics[width=0.40\textwidth]{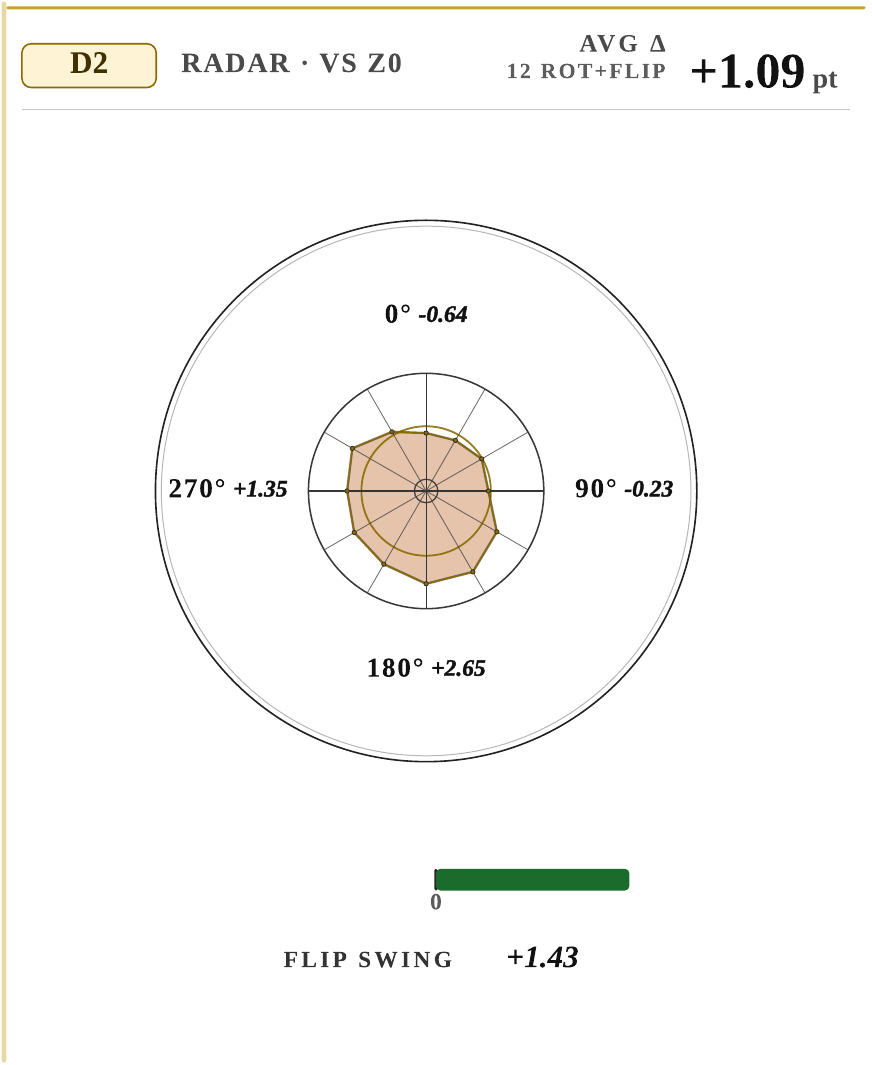} \\
      \includegraphics[width=0.40\textwidth]{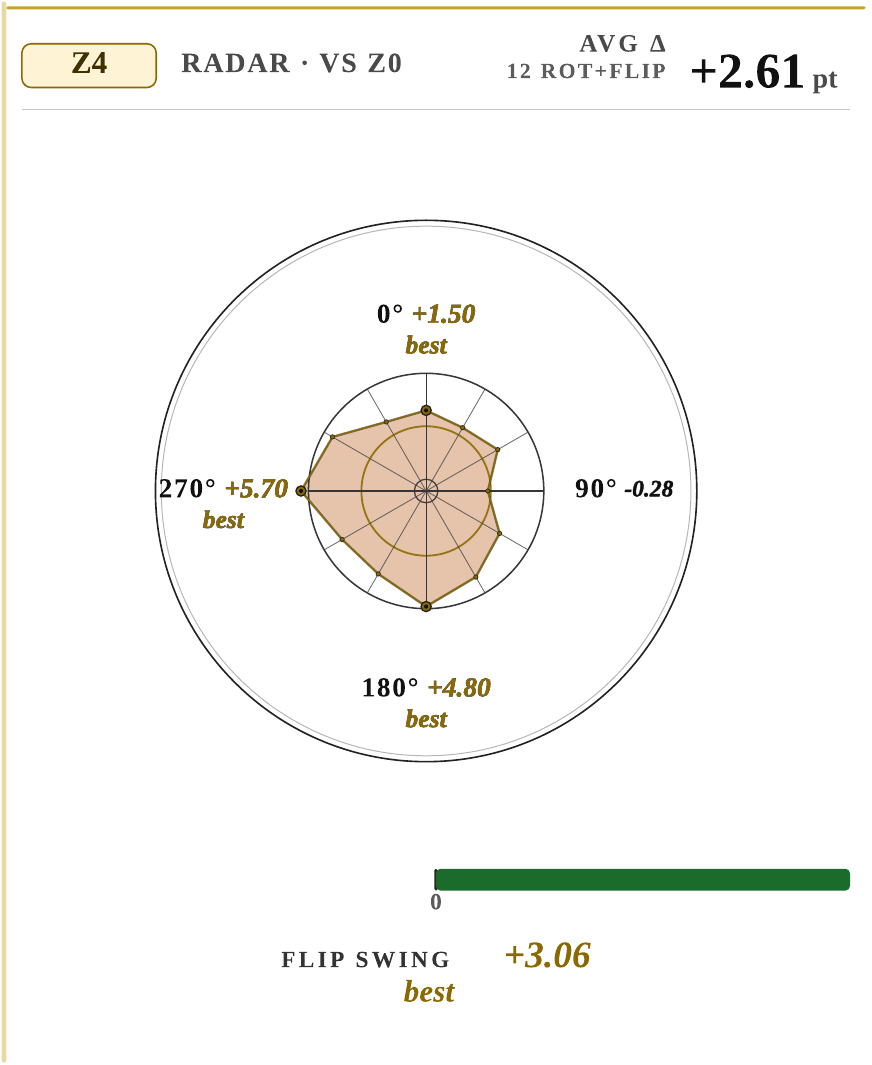} &
      \includegraphics[width=0.40\textwidth]{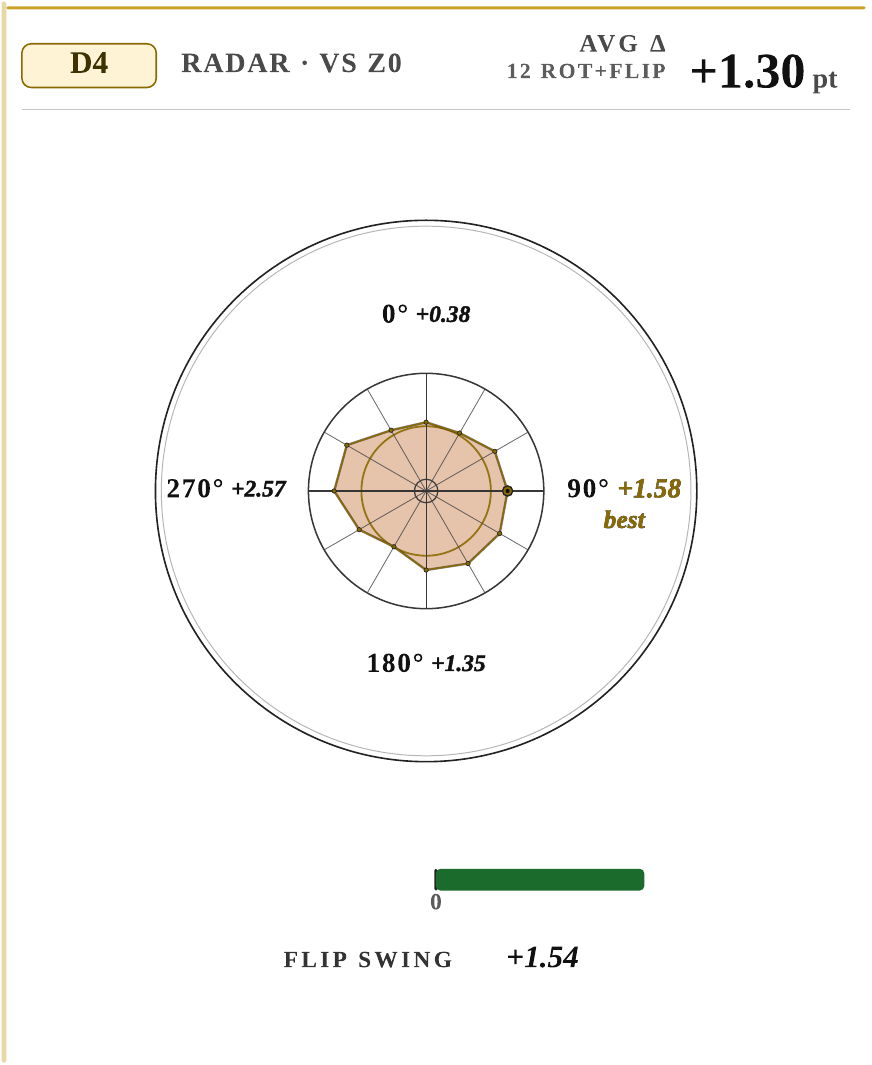}
  \end{tabular}}
  \endgroup
  \caption{Orbit-specific transformation-wise gains over the no-generator $C_0$ control. Positive values identify where exposed cyclic or dihedral generators improve the sampled $\mathrm{O}(2)$ orbit; the reader should prioritize these transformed-orbit gains rather than the clean control.}
  \label{fig:mnist-o2-z0-panels}
\end{figure}

Finally, the orbit-flattening diagnostic is directionally consistent but should be interpreted conservatively. The standard deviation of seed-mean accuracies across the thirteen transformations decreases from 24.88 for $C_0$ to 24.71 for $C_2$, 24.10 for $C_4$, 23.98 for $D_2$, and 24.64 for $D_4$. Thus all generator-aware variants produce a slightly flatter orbit profile than the no-generator control. However, the absolute spread is dominated by the intrinsic difficulty gap between near-identity rotations, which remain near $95\%$, and far rotations, which are near $40\%$. We therefore use orbit flatness only as a secondary behavioral diagnostic; the primary evidence is the controlled improvement in held-out-orbit average and worst-case transformed accuracy. Fig.~\ref{fig:mnist-o2-z0-panels} reports generator-aware variants relative to the no-generator $C_0$ control, so positive values indicate gains from exposing cyclic or dihedral generator streams within the same GARI-Net backbone.

\subsection{\texorpdfstring{Controlled $\mathrm{SO}(3)$ Generator-Decomposition Probe}{Controlled SO(3) Generator-Decomposition Probe}}
ModelNet40 is used here as a controlled $\mathrm{SO}(3)$ generator-accessibility diagnostic, not as a claim of state-of-the-art rotation-invariant 3D recognition. Dedicated 3D equivariant or invariant architectures can encode stronger geometric priors through irreducible representations, tensor products, spherical harmonics, local frames, or invariant point-cloud operators. Our question is different: if only $X/Y$ axial generator interfaces are exposed during training, does the model improve on a held-out $Z$-axis generator, and does the representation-level DEE move consistently with that transfer?

We use $\mathrm{SO}(3)$ as a controlled generator-accessibility probe rather than as a control-theoretic controllability claim. The axial decomposition of three-dimensional rotations is a standard structural fact; our question is whether a generic sequence backbone equipped with an GARI-Net generator-aligned interface can operationalize this structure as transferable soft robustness. We therefore expose generator structure only along the $X$ and $Y$ axes, leave the $Z$-axis family unaugmented and without an explicit generator, and ask whether strengthening the $X/Y$ interface improves held-out $Z$-axis rotation generalization. The $\mathrm{SO}(3)$ experiment is therefore a neural accessibility diagnostic: it asks whether exposing two axial generator directions through GARI-Net makes the learned sequence representation more useful on a structurally related but held-out axial family.

This design is motivated by the generator structure of $\mathrm{SO}(3)$: axial rotations provide a natural decomposition of three-dimensional rotations, and compositions of rotations around different axes expose the non-commutative structure absent from planar cyclic tests. The $X/Y$-to-$Z$ protocol is also motivated by the local Lie-algebra structure of $\mathrm{SO}(3)$: infinitesimal rotations around $X$ and $Y$ generate the $Z$ direction through their Lie bracket. This gives a principled local reason to test whether a stronger $X/Y$ generator interface transfers to the held-out $Z$ family, and it is a concrete instance of the generator-family reachability principle in Appendix~\ref{app:separable-lie-soft-equivariance}: the exposed $X/Y$ infinitesimal directions bracket-generate the missing $Z$ Lie-algebra direction at the group-geometry level. The interpretation remains local and diagnostic; finite large-angle $X/Y$ probes do not certify arbitrary $\mathrm{SO}(3)$ equivariance. We do not claim this decomposition as a contribution. Instead, it gives a structured setting in which the exposed generator directions and the held-out generator family can be separated by design. The finite-order evaluation therefore tests whether stronger implemented $X/Y$ generator structure can transfer to the deliberately withheld $Z$ family inside an GARI-Net architecture, not whether the paper proves a new geometric controllability or reachability theorem for $\mathrm{SO}(3)$.

\paragraph{Claim--control--endpoint.}
The claim tested here is whether strengthening the exposed $X/Y$ axial generator interface improves robustness on a structurally related but held-out $Z$-axis rotation family. $C_0$ removes the $X/Y$ generator interface and serves as a no-generator lower/control reference. The $X/Y/Z$ $n=4$ setting explicitly exposes the $Z$ generator and is included as an explicit-generator reference, not as held-out-transfer evidence. The primary endpoint is $Z$ gen Top-1 over $30^\circ$--$330^\circ$, excluding $0^\circ$; clean $0^\circ$ and $X/Y$ gen are controls. The claim would be weakened if $X/Y$ $n=4$ improved mainly at clean $0^\circ$, if $C_0$ matched $X/Y$ $n=4$ on $Z$ gen, or if $X/Y/Z$ $n=4$ were the only setting that improved $Z$-axis performance. Observationally, $X/Y$ $n=4$ improves $Z$ gen by $+8.97$ percentage points over $X/Y$ $n=2$ while clean $0^\circ$ improves only $+0.33$ percentage points. $C_0$ is lower at $26.80\%$, supporting that the generator interface matters. The $X/Y/Z$ $n=4$ reference gives only $+0.51$ percentage points on $Z$ gen over $X/Y$ $n=4$ under the same no-$Z$-augmentation protocol, suggesting that the $X/Y$-only interface already captures much of the $Z$-axis task robustness, while explicit $Z$ structure primarily improves the representation diagnostic.

\paragraph{Experimental setup.}
We evaluate on ModelNet40 point-cloud classification using 2048 input points per object. The ModelNet40 settings reported here are architecture-matched rather than recipe-locked: $C_0$, $X/Y$ $n=2$, $X/Y$ $n=4$, and $X/Y/Z$ $n=4$ use the same 3D GARI-Net backbone family, voxelization pipeline, embedding dimension 256, depth 1, training data, augmentation schedule, evaluation script, seed set, parameterization, and parameter scale of about 8.30M. To obtain stronger equivariant learning and representation-extraction capacity, the shared \texttt{AutoregressiveBlock} is instantiated by Mamba2 in all compared models. The primary architectural intervention is the exposed generator-interface strength. The weak setting, $n=2$, applies $180^\circ$ generators around the $X$ and $Y$ axes, implementing $C_2\times C_2$. The strong setting, $n=4$, applies $90^\circ$ generators around the same two axes, implementing $C_4\times C_4$. In both primary settings, the $Z$ axis is intentionally left without a generator. Standard optimization and regularization hyperparameters are selected within each setting using validation performance rather than forced to be identical. Thus the result should be read as a single-factor architectural intervention under setting-specific validated training, not as a fully locked-recipe ablation; this prevents any setting from being artificially handicapped, and is why the clean $0^\circ$ score is treated only as a sanity check rather than as evidence for rotation generalization. Training augmentation is also intentionally asymmetric across all settings: random rotations are applied only around $X$ and $Y$, within $\pm60^\circ$, and no training rotation is applied around $Z$. Thus any improvement on nontrivial $Z$-axis test rotations cannot be attributed to direct $Z$-axis architectural supervision in the primary $X/Y$ settings or to memorizing $Z$-axis augmentation.

Two additional controls are completed for the extended ModelNet40 protocol. First, a $C_0$ no-generator control uses the same 3D GARI-Net pipeline but no $X/Y/Z$ equivariant PatchEmbed generator; it is the lower reference for excluding explanations based only on the 3D pipeline, capacity, training schedule, or data augmentation. Second, an $X/Y/Z$ $n=4$ setting exposes explicit $C_4$ generator structure on all three axes and serves as an explicit-generator reference for the $Z$ family. It is not held-out transfer evidence, because the $Z$ generator is directly exposed. Since it is trained without $Z$-axis rotation augmentation, it is not a $Z$-augmentation upper bound.

Accordingly, the ModelNet40 comparison is kept within the same 3D GARI-Net pipeline and varies generator exposure under the same no-$Z$-augmentation protocol. PointNet-style, graph-based, voxel, or explicitly rotation-robust methods remain important SOTA context, but they differ in point representation, neighborhood construction, voxelization, pooling, and augmentation policy, so they would test complete-system competitiveness rather than the generator-interface mechanism isolated here. The $X/Y/Z$ $n=4$ setting is therefore used only as an explicit-generator reference, not as the held-out transfer result or as a $Z$-augmentation upper bound. The conclusion is correspondingly scoped: this experiment supports controlled $\mathrm{SO}(3)$ decomposition evidence for held-out generator transfer within the 3D GARI-Net setting, not an absolute ModelNet40 leaderboard claim.

\paragraph{Evaluation protocol.}
For each seed-level checkpoint, we evaluate the clean test set after applying a fixed rotation around exactly one axis at a time. Each of the three axes, $X$, $Y$, and $Z$, is evaluated at twelve angles, $0^\circ,30^\circ,\ldots,330^\circ$. We report four aggregation views. The $0^\circ$ score averages the no-rotation evaluation across the three axis-specific test loaders and serves only as a clean-orientation sanity check. The all-angle score averages all twelve angles for a given axis. The $X/Y$ generalization score excludes the training-supported interval, $|\theta|\leq60^\circ$, and averages the remaining seven angles from $90^\circ$ to $270^\circ$. The $Z$ generalization score excludes only $0^\circ$ and averages the eleven nontrivial $Z$ rotations from $30^\circ$ to $330^\circ$, because the training distribution contains no $Z$ rotations at all. We use this $Z$ generalization metric as the primary readout. For the rotation-generalization summaries, the reported standard deviations are computed after averaging over seeds at each fixed transformation and therefore mainly reflect heterogeneity across rotation angles, whose induced test distributions have different intrinsic difficulty. Increasing the number of seeds would reduce only the seed-estimation component, not this angle-distribution component; we therefore prioritize a paired three-seed protocol together with angle-resolved gap plots. This design also avoids treating the twelve angles as independent replicates for a formal significance test: the angles are structured interventions generated by the same rotation group, and their role is to localize the effect relative to the implemented and held-out generators.

The axial protocol is not a proxy chosen for convenience. Because general three-dimensional rotations can be decomposed into coordinate-axis rotations, evaluating the $X$-, $Y$-, and $Z$-axis families separately gives a generator-resolved decomposition of the $\mathrm{SO}(3)$ behavior. This design is more diagnostic than a single random-rotation aggregate for the claim tested here, because a random aggregate would mix the implemented $X/Y$ components with the held-out $Z$ component into one scalar and obscure whether the held-out $Z$ component improves when only the $X/Y$ generator interface is strengthened. At the same time, we do not claim that this single-axis protocol exhausts all numerical effects that may arise under arbitrary composed rotations. Since $\mathrm{SO}(3)$ is non-commutative, composition order, voxelization, interpolation, sampling, and error accumulation can create distributions that are not identical to any one-axis family. The experiment therefore tests held-out generator transfer in a controlled and identifiable form, rather than claiming full robustness over random composed rotations.

\begin{table}[htbp]
  \centering
  \small
  \setlength{\tabcolsep}{4pt}
  \renewcommand{\arraystretch}{1.10}
  \begin{tabularx}{\textwidth}{%
      >{\raggedright\arraybackslash}p{0.34\textwidth}
      >{\centering\arraybackslash}p{0.11\textwidth}
      >{\centering\arraybackslash}p{0.08\textwidth}
      >{\centering\arraybackslash}p{0.12\textwidth}
    >{\raggedright\arraybackslash}X}
    \toprule
    Setting & Clean $0^\circ$ & $Z$ gen & $Z$ all-angle & Interpretation \\
    \midrule
    $C_0$ (no generator) & 89.59 & 26.80 & 32.03 & No-generator control \\
    $X/Y$ $n=2$ & 90.19 & 29.19 & 34.27 & Weak X/Y generator \\
    $X/Y$ $n=4$ & 90.52 & 38.16 & 42.52 & Primary X/Y transfer \\
    $X/Y/Z$ $n=4$ & 90.15 & 38.67 & 42.96 & Explicit Z reference \\
    \midrule
    $\Delta$ $X/Y$ $n=4 - X/Y$ $n=2$ & +0.33 & \best{+8.97} & \best{+8.25} & Held-out Z gain \\
    $\Delta$ $X/Y/Z$ $n=4 - X/Y$ $n=4$ & -0.37 & +0.51 & +0.44 & Small explicit-Z gain \\
    \bottomrule
  \end{tabularx}
  \caption{Primary ModelNet40 fixed-axis rotation results under controlled generator exposure. Values are Top-1 means in percent. All settings use the same ModelNet40 protocol and no $Z$-axis rotation augmentation. $X/Y/Z$ $n=4$ exposes the $Z$ generator structurally but is not trained with $Z$-axis rotation augmentation; therefore it should not be read as a $Z$-augmentation upper bound.}
  \label{tab:modelnet40-so3-mean}
\end{table}

\paragraph{Main results.}
\Cref{tab:modelnet40-so3-mean} gives the primary ModelNet40 $Z$-axis results and key attribution deltas under the completed $C_0$, $X/Y$ $n=2$, $X/Y$ $n=4$, and $X/Y/Z$ $n=4$ settings. Full angle-wise ModelNet40 rotation gaps, $X/Y$-axis controls, and extended controls are reported in the supplementary material. The main signal appears on the held-out component targeted by the protocol. Strengthening the $X/Y$ interface from $n=2$ to $n=4$ raises held-out $Z$ generalization from $29.19\%$ to $38.16\%$, a gain of $+8.97$ percentage points, while the clean $0^\circ$ score changes by only $+0.33$ percentage points. $C_0$ is lower at $26.80\%$ on $Z$ gen, supporting that the generator interface matters. The $X/Y/Z$ $n=4$ explicit-generator reference reaches $38.67\%$ on $Z$ gen, only $+0.51$ percentage points above $X/Y$ $n=4$ under the same no-$Z$-augmentation protocol.

The completed controls sharpen the attribution without changing the primary claim. $C_0$ is a no-generator lower/control reference, while $X/Y/Z$ $n=4$ provides an explicit-generator reference. Since $X/Y/Z$ $n=4$ is trained without $Z$-axis rotation augmentation, it is not a $Z$-augmentation upper bound and should not be counted as held-out transfer evidence. Under this protocol, a separate $Z$-axis generator is not necessary to obtain the observed held-out $Z$-axis gain from $X/Y$ $n=2$ to $X/Y$ $n=4$; this does not imply that $Z$-axis structure or $Z$-axis augmentation is unnecessary in general.

The $X$ and $Y$ axes serve primarily as controls for the primary $X/Y$ $n=4$ versus $X/Y$ $n=2$ comparison. Their generalization gains are small, $+0.14$ and $+1.43$ percentage points, respectively, and the clean score changes by only $+0.33$ percentage points. If the result were merely a generic accuracy improvement from the $n=4$ model, one would expect a comparable shift at $0^\circ$ or a more uniform improvement across axes. Instead, the largest change occurs on the axis family that receives no direct equivariant structure in the primary comparison. We interpret this as generator-resolved soft transfer: the model does not merely improve the axes it directly sees, but benefits on a held-out axial family that is structurally related through the $\mathrm{SO}(3)$ decomposition. The result does not establish exact $\mathrm{SO}(3)$ equivariance or full robustness to arbitrary composed rotations.

To complement this task-level result with a direct feature-level diagnostic, we also compute Direct Equivariance Error (DEE) on frozen ModelNet40 checkpoints. Let $h_\ell(x)$ denote the final post-fusion pre-pooling token/voxel representation at layer $\ell$, let $\mathcal{N}$ be channel-wise layer normalization, and let $\rho_{z,\alpha}$ be the permutation induced on the voxel/token grid by the $Z$-axis rotation $R_z(\alpha)$. Operationally, $\tau_g$ is this known voxel/token reindexing for each held-out $Z$ rotation; no fitted representation action is used. For each evaluated held-out $Z$ rotation, we compute
\begin{equation}
  \label{eq:modelnet40-so3-z-dee}
  \mathrm{DEE}^{(\ell)}(R_z(\alpha))
  =
  \mathbb{E}_{x}
  \frac{
    \left\lVert
    \mathcal{N}\!\left(h_\ell(R_z(\alpha)x)\right)
    -
    \rho_{z,\alpha}\mathcal{N}\!\left(h_\ell(x)\right)
    \right\rVert_F
  }{
    \frac{1}{2}\left(
      \left\lVert\mathcal{N}\!\left(h_\ell(R_z(\alpha)x)\right)\right\rVert_F
      +
      \left\lVert\mathcal{N}\!\left(h_\ell(x)\right)\right\rVert_F
    \right)
    +\varepsilon
  }.
\end{equation}
As in the ImageNet-1K and MNIST diagnostics, DEE is not used as a training objective; it is a post-hoc diagnostic of final-representation consistency. We report the single half-turn error and the quarter-turn orbit average,
\begin{equation}
  \label{eq:modelnet40-so3-z-dee-summary}
  \begin{aligned}
    \mathrm{DEE}_{Z\text{-}180} &= \mathrm{DEE}(R_z(180^\circ)),\\
    \mathrm{DEE}_{Z\text{-}C_4}
    &= \frac{1}{3}\bigl[
      \mathrm{DEE}(R_z(90^\circ))\\
      &\quad +\mathrm{DEE}(R_z(180^\circ))
      +\mathrm{DEE}(R_z(270^\circ))
    \bigr].
  \end{aligned}
\end{equation}

\begin{table}[tbp]
  \centering
  \caption{Compact post-hoc $Z$-axis DEE and held-out $Z$ accuracy on ModelNet40. DEE is a conservative representation diagnostic computed on frozen checkpoints and is not used as a training objective. The primary endpoint remains held-out $Z$ classification accuracy. Top-1 values are means in percent. Angle-wise DEE values are provided in the supplementary material.}
  \label{tab:modelnet40-so3-z-dee}
  \scriptsize
  \setlength{\tabcolsep}{4pt}
  \renewcommand{\arraystretch}{1.10}
  \begin{tabularx}{\textwidth}{%
      >{\raggedright\arraybackslash}p{0.38\textwidth}
      >{\centering\arraybackslash}X
      >{\centering\arraybackslash}X
    >{\centering\arraybackslash}X}
    \toprule
    Setting & $\mathrm{DEE}_{Z\text{-}C_4}\downarrow$ & $Z$ gen Top-1$\uparrow$ & Clean $0^\circ$\,$\uparrow$ \\
    \midrule
    $C_0$ / no generator & 1.121 & 26.80 & 89.59 \\
    $X/Y$ $n=2$ & 1.141 & 29.19 & 90.19 \\
    $X/Y$ $n=4$ & 1.136 & 38.16 & 90.52 \\
    $X/Y/Z$ $n=4$ & 1.095 & 38.67 & 90.15 \\
    $\Delta$ $X/Y$ $n=4 - X/Y$ $n=2$ & -0.005 & +8.97 & +0.33 \\
    $\Delta$ $X/Y/Z$ $n=4 - X/Y$ $n=4$ & -0.041 & +0.51 & -0.37 \\
    \bottomrule
  \end{tabularx}
\end{table}

Full angle-wise ModelNet40 rotation gaps, angle-wise $Z$-axis DEE, and extended controls are reported in the supplementary material.

\Cref{tab:modelnet40-so3-z-dee} shows that the primary task-level conclusion is stronger than the representation-diagnostic change in the $X/Y$-only comparison. Moving from $X/Y$ $n=2$ to $X/Y$ $n=4$ raises the held-out $Z$ generator Top-1 from $29.19\%$ to $38.16\%$, a gain of $+8.97$ percentage points, and raises the $Z$ all-angle Top-1 from $34.27\%$ to $42.52\%$, a gain of $+8.25$ percentage points. By contrast, the clean $0^\circ$ score increases by only $+0.33$ percentage points, and $\mathrm{DEE}_{Z\text{-}C_4}$ changes only slightly from $1.141$ to $1.136$.

The explicit-generator reference separates representation-level alignment from task-level robustness. $X/Y/Z$ $n=4$ lowers $\mathrm{DEE}_{Z\text{-}C_4}$ substantially relative to $X/Y$-only $n=4$, from $1.136$ to $1.095$, but $Z$ gen improves only from $38.16\%$ to $38.67\%$. Thus explicit $Z$ generator exposure improves the frozen representation diagnostic, but without $Z$-axis rotation augmentation the classifier does not convert this into a large additional $Z$-axis accuracy gain.

The supplementary angle-wise diagnostic reinforces this distinction. In the primary $X/Y$ comparison, the $n=4$ setting lowers DEE at $90^\circ$ and $270^\circ$ by $0.045$ and $0.040$, respectively, but has a higher DEE at $180^\circ$ by $0.071$, producing only a $-0.005$ change in $\mathrm{DEE}_{Z\text{-}C_4}$. In the explicit-generator reference comparison, $X/Y/Z$ $n=4$ lowers all three quarter-turn DEE values relative to $X/Y$ $n=4$, with $\mathrm{DEE}_{Z\text{-}C_4}$ decreasing by $0.041$. We therefore use DEE only as an accompanying representation diagnostic; the primary endpoint remains held-out $Z$ classification accuracy.

\paragraph{Scope of the claim.}
The supported claim is not that two axes are sufficient for exact $\mathrm{SO}(3)$ equivariance, nor that a $Z$-axis generator would be unnecessary in general. The supported claim is narrower and more diagnostic: under a controlled ModelNet40 protocol with no $Z$-axis augmentation and no $Z$-axis PatchEmbed generator in the primary setting, strengthening the exposed $X/Y$ generator interface produces a large improvement on the held-out $Z$ generator family. Under this protocol, a separate $Z$-axis generator is not necessary to obtain the observed held-out $Z$-axis gain from $X/Y$ $n=2$ to $X/Y$ $n=4$; this does not imply that $Z$-axis structure or $Z$-axis augmentation is unnecessary in general. The $C_0$ and $X/Y/Z$ $n=4$ controls serve as lower/control and explicit-generator reference checks for this interpretation, not new leaderboard comparisons. The result still does not verify classical controllability, certify full $\mathrm{SO}(3)$ equivariance, guarantee robustness to arbitrary composed rotations, or replace an explicit three-axis equivariant design.

\subsection{Cross-Experiment Synthesis}
The experimental chain should be read as a sequence of questions rather than as interchangeable robustness benchmarks. First, does ordinary supervision expose an internal symmetry-alignment signal? The ImageNet-1K loss-dynamics diagnostic shows that a sequence-level symmetry signal can emerge without backpropagating an explicit symmetry loss. Second, does the sequence-side premise hold? The GenomicBenchmarks reversal diagnostic tests whether task-relevant evidence is preserved under a generator-transformed input, without claiming ordinary genomic SOTA. Third, does the generator-aligned interface scale to large visual data? The ImageNet-1K rotation experiment serves as the large-scale anchor and pairs orbit-localized accuracy with frozen-checkpoint DEE. Fourth, does generator coverage matter under a controlled planar backbone? The MNIST $\mathrm{O}(2)$ experiment isolates cyclic and dihedral generator exposure, including the $C_0$ no-generator control. Fifth, does generator structure transfer to a held-out non-planar family? The ModelNet40 study tests $C_0$, $X/Y$ $n=2$, $X/Y$ $n=4$, and an explicit-generator $X/Y/Z$ $n=4$ reference for the $Z$ family without counting that reference as held-out transfer evidence.

Taken together, the experiments presented support the diagnostic form of the generator\hyp{}aligned soft\hyp{}equivariance claim across sequence, planar visual, and three-dimensional settings. The ImageNet results support large-scale held-out rotation generalization, GenomicBenchmarks supports the sequence-reversal premise, MNIST $\mathrm{O}(2)$ isolates generator coverage under controlled conditions, and ModelNet40 supports held-out generator transfer under a controlled $\mathrm{SO}(3)$ axial decomposition. The added $C_0$ and $X/Y/Z$ $n=4$ controls sharpen the ModelNet40 interpretation: $C_0$ shows that removing the generator interface weakens held-out $Z$ robustness, while $X/Y/Z$ $n=4$ shows that explicitly exposing the $Z$ generator mainly improves the representation-level DEE diagnostic and only marginally improves $Z$ gen accuracy without $Z$-axis augmentation. Thus the main result is not a leaderboard claim, but a localized generator-transfer result under controlled exposure and training coverage.


The common message is therefore not that GARI-Net enforces exact equivariance by hard architectural constraints, nor that it replaces the many orthogonal tools used to raise clean or augmentation-supported accuracy, such as stronger backbones, pretraining, distillation, longer training, or generic data augmentation~\cite{He2016ResNet,Liu2021Swin,He2022MAE,Hinton2015Distilling,Zhang2018Mixup,Yun2019CutMix,Cubuk2020RandAugment,Zhong2020RandomErasing}. Instead, GARI-Net provides a generator-aligned learning interface through which equivariant behavior can emerge from task supervision, be strengthened by generator coverage, and extend from sequence symmetry to planar and three-dimensional held-out geometric generalization. This is the soft-equivariance interpretation supported by the experiments: the architecture does not prescribe the full group action in advance, but makes the relevant group structure learnable and empirically testable across increasingly expressive symmetry settings. Its value should therefore be judged primarily on generator-localized and out-of-training-distribution transformations, with all-angle averages treated only as diluted controls. The synthesis supports diagnostic soft equivariance under specified probes and controls, not exact continuous-group certification.

\section{Conclusion}

This paper developed generator-aligned representation interfaces as a route to diagnostic soft equivariance under Lie group actions. The central idea is to move group structure from a fully hard-coded equivariant operator into an interface that a generic sequence backbone can use: exposed generators instantiate canonical and generator-indexed streams, frame conversion aligns processing and interaction coordinates, cross-stream interaction shares evidence across views, and discrepancy-aware fusion avoids blind terminal averaging. In this sense, GARI-Net is not only a model variant, but an implementation of a broader generator-indexed interface principle.

The experiments support this principle across sequence, planar visual, and three\hyp{}dimensional settings. ImageNet-1K provides the large-scale anchor by testing held-out rotation generalization beyond the directly supported region. GenomicBenchmarks tests whether the sequence-reversal premise carries useful transformed evidence. MNIST $\mathrm{O}(2)$ isolates generator coverage under controlled conditions and shows why DEE must be interpreted together with discriminative transformed accuracy. ModelNet40 tests held-out generator transfer under an $\mathrm{SO}(3)$ axial decomposition; the $C_0$ and $X/Y/Z$ $n=4$ controls sharpen the interpretation by separating no-generator behavior, $X/Y$ transfer, and explicit $Z$-generator exposure. Across these experiments, DEE is used as a frozen-checkpoint representation diagnostic rather than as a training objective or a complete explanation of task accuracy.

The scope of the claim is intentionally bounded. Finite generator probes do not certify exact continuous-group equivariance. Held-out robustness is not identical to equivariance. Low DEE is not sufficient evidence without task-level behavior. The ModelNet40 protocol is a controlled generator-transfer probe, not a proof of full $\mathrm{SO}(3)$ robustness under arbitrary composed rotations. These limitations are not incidental caveats; they define the diagnostic interpretation of the paper.

The significance of GARI lies in shifting the unit of architectural design from group-specific layers to generator-aligned interfaces surrounding generic backbones, rather than in seeking comprehensive performance dominance over specialized equivariant systems. This formulation is particularly relevant when exact-equivariant operators are unavailable, costly to redesign, or difficult to integrate with contemporary sequence architectures. Accordingly, the pertinent evaluation criterion is whether the proposed interface provides a reusable and empirically measurable mechanism for inducing soft equivariance across data modalities, rather than whether it surpasses every specialized model on its respective benchmark. The diagnostics provide evidence for this cross-modal utility, while the matched Tiny-ImageNet protocol is intended to complete the component-level architectural attribution.



\appendix
\section{Additional Theory and Proofs}
\label{app:theory-proofs}
\label{app:separable-lie-soft-equivariance}
This appendix keeps only the proof details needed for the main paper to be self-contained. First, dense exactness is an ideal boundary case: if $H\subseteq G$ is dense, $g\mapsto\alpha_gx$, $\tau_g$, and $F$ are continuous, and $F(\alpha_hx)=\tau_hF(x)$ for every $h\in H$, then the same identity holds for every $g\in G$ by taking $h_n\to g$ and passing to the limit. Finite experiments do not satisfy this hypothesis; they only evaluate a declared probe set.

For a compact evaluation region $K\subseteq G$, a finite $\varepsilon$-net $P\subset K$, and pointwise residual $r_F(x,g)=d_{\mathcal{Y}}(F(\alpha_gx),\tau_gF(x))$, define $\omega_x(\varepsilon)=\sup_{d_G(g,h)\leq\varepsilon}|r_F(x,g)-r_F(x,h)|$. Then every $g\in K$ has $p\in P$ with
\begin{equation}
  r_F(x,g)\leq \max_{p\in P} r_F(x,p)+\omega_x(\varepsilon).
  \label{eq:app-finite-net-bound}
\end{equation}
Thus finite-probe DEE is meaningful only relative to coverage and local stability assumptions; it is not a continuous-group certificate.

For generator-family reachability, let $\mathcal{A}=\{A_1,\ldots,A_q\}\subset\mathfrak{g}$ be the exposed Lie-algebra directions. If the Lie closure $\operatorname{Lie}(\mathcal{A})$ equals the Lie algebra of the identity component, compositions of the one-parameter flows $\exp(tA_i)$ generate that component at the group-geometry level. GARI-Net uses this fact only to choose meaningful exposed and held-out generator families, not as a neural controllability theorem.

For the implemented generator-indexed map with exposed set $\mathcal{S}=\{s_1,\ldots,s_q\}$, $z_{\mathcal{S}}(x)=(P_\theta(x),\rho_{s_1}P_\theta(x),\ldots,\rho_{s_q}P_\theta(x))$ and $F_{\mathcal{S}}(x)=\Omega_\omega(\Psi_\omega(z_{\mathcal{S}}(x)))$. Inserting and subtracting the two intermediate product-space terms gives the residual localization
\begin{align}
  d_{\mathcal{Y}}(F_{\mathcal{S}}(\alpha_gx),\tau_gF_{\mathcal{S}}(x))
  &\leq L_\Omega L_\Psi d_{\times}(z_{\mathcal{S}}(\alpha_gx),\mathcal{T}_gz_{\mathcal{S}}(x)) \notag\\
  &\quad +L_\Omega d'_{\times}(\Psi_\omega(\mathcal{T}_gz_{\mathcal{S}}(x)),\mathcal{T}'_g\Psi_\omega(z_{\mathcal{S}}(x))) \notag\\
  &\quad +d_{\mathcal{Y}}(\Omega_\omega(\mathcal{T}'_g\Psi_\omega(z_{\mathcal{S}}(x))),\tau_g\Omega_\omega(\Psi_\omega(z_{\mathcal{S}}(x)))).
  \label{eq:app-product-space-three-term}
\end{align}
The terms are interface mismatch, stream-compatibility mismatch, and terminal-fusion mismatch. A standard telescoping argument further bounds a stack by downstream Lipschitz constants times layerwise commutation errors. For reversal, a causal local window does not commute with sequence reversal, while symmetric or bidirectional context preserves the local neighborhood away from boundaries; this motivates ODBC as context repair rather than exact equivariance.

\section{Core Diagnostic Derivations and Residual Details}
\label{app:core-diagnostic-derivations}
\label{app:symmetry-loss-diagnostic}
For the monitored SSM-style diagnostic, a discretized state update gives $h_t=\bar{\mathbf{A}}h_{t-1}+\bar{\mathbf{B}}x_t$ and $y_t=\mathbf{C}h_t$, so the previous-input influence is $I_t=\partial y_t/\partial x_{t-1}=\mathbf{C}\bar{\mathbf{A}}\bar{\mathbf{B}}$. Under zero-order-hold discretization and a first-order small-step approximation, $I_t\approx\Delta\mathbf{C}\mathbf{B}$. In Mamba-style blocks we monitor the streamwise projection proxy $I^{(l)}=\mathbf{W}_{C}^{(l)\top}\mathbf{W}_{B}^{(l)}$ and
\begin{equation}
  \ell_{sym}=\sum_{l=1}^{L}\bar{\alpha}_l\left\|I^{(l,\mathrm{id})}-I^{(l,\mathrm{gen})}\right\|.
  \label{eq:app-symmetry-loss}
\end{equation}
In the ImageNet loss-dynamics experiment the weights $\bar{\alpha}_l$ are fixed and this quantity is not backpropagated. DEE is likewise interpreted as a frozen-checkpoint representation diagnostic on pre-specified transformations; low DEE is informative only when paired with task accuracy because smooth or non-discriminative representations can also have small residuals.


\vskip 0.2in
\bibliography{gari}

\end{document}